\documentclass[11pt]{article}

\usepackage[final]{acl}

\usepackage{times}
\usepackage{latexsym}
\usepackage{amsmath}
\usepackage{booktabs}
\usepackage{graphicx}
\usepackage{newfloat}
\usepackage{array}
\usepackage{multirow}
\usepackage{xcolor}
\usepackage{enumitem}
\usepackage{listings}
\usepackage[T1]{fontenc}
\usepackage[utf8]{inputenc}
\usepackage{microtype}
\usepackage{inconsolata}

\DeclareFloatingEnvironment[fileext=loa,name=Algorithm]{algorithm}

\newcommand{\method}{\textsc{MAIGO}}
\newcommand{\qwenSeven}{Qwen2.5-7B-Instruct}
\newcommand{\qwenThree}{Qwen2.5-3B-Instruct}
\newcommand{\deepseekvthree}{DeepSeek-V3.1-Terminus}

\title{MAIGO: Mitigating Lost-in-Conversation with History-Cleaned On-Policy Self-Distillation}

\author{
Haoyu Zheng\textsuperscript{1,5}\thanks{Work done during internship at Tencent.},
Yun Zhu\textsuperscript{2},
Shu Yuan\textsuperscript{3},
Shangming Chen\textsuperscript{4},
Qing Wang\textsuperscript{5},\\
Wenqiao Zhang\textsuperscript{1}\thanks{Corresponding author.},
Jun Xiao\textsuperscript{1},
Yueting Zhuang\textsuperscript{1}
\\
\textsuperscript{1}Zhejiang University
\quad \textsuperscript{2}Ant Group
\\
\textsuperscript{3}Huazhong University of Science and Technology
\quad
\textsuperscript{4}Fuzhou University
\\
\textsuperscript{5}Tencent
}

\begin{document}
\maketitle

\begin{abstract}

Large language models often solve tasks from a fully specified prompt but
degrade when the same requirements unfold over multiple turns, known as the
lost-in-conversation (LiC) gap. We trace part of this degradation to
self-contamination: intermediate assistant replies enter later context and
carry early deviations forward. Motivated
by this mechanism, we propose \method{}\footnote{Code is available at
\url{https://github.com/Chihaya-Anon-chan/MAIGO}.}, an on-policy
self-distillation method that reduces this contamination using history-cleaned
references from the model's own policy. For middle turns, \method{} removes
prior assistant replies while preserving the user-visible sharded prefix; for
answer turns, it distills from paired full-view references conditioned on the
completed user-side dialogue. A reliability weight downweights middle-turn
samples that disagree with the clean reference. \method{} requires no verifier
rewards, state labels, or inference-time scaffolding. Under the LiC paired-view
protocol with deterministic verifiers, \method{} improves \qwenSeven{}
SHARDED accuracy from 52.8 to 66.1 and the SHARDED/FULL ratio from 66.5\% to
84.1\%, while keeping FULL accuracy within 2.3 points. These results show that self-contamination is a trainable component of the LiC gap.

\end{abstract}

\section{Introduction}

\begin{figure}[t]
    \centering
    \includegraphics[width=\columnwidth]{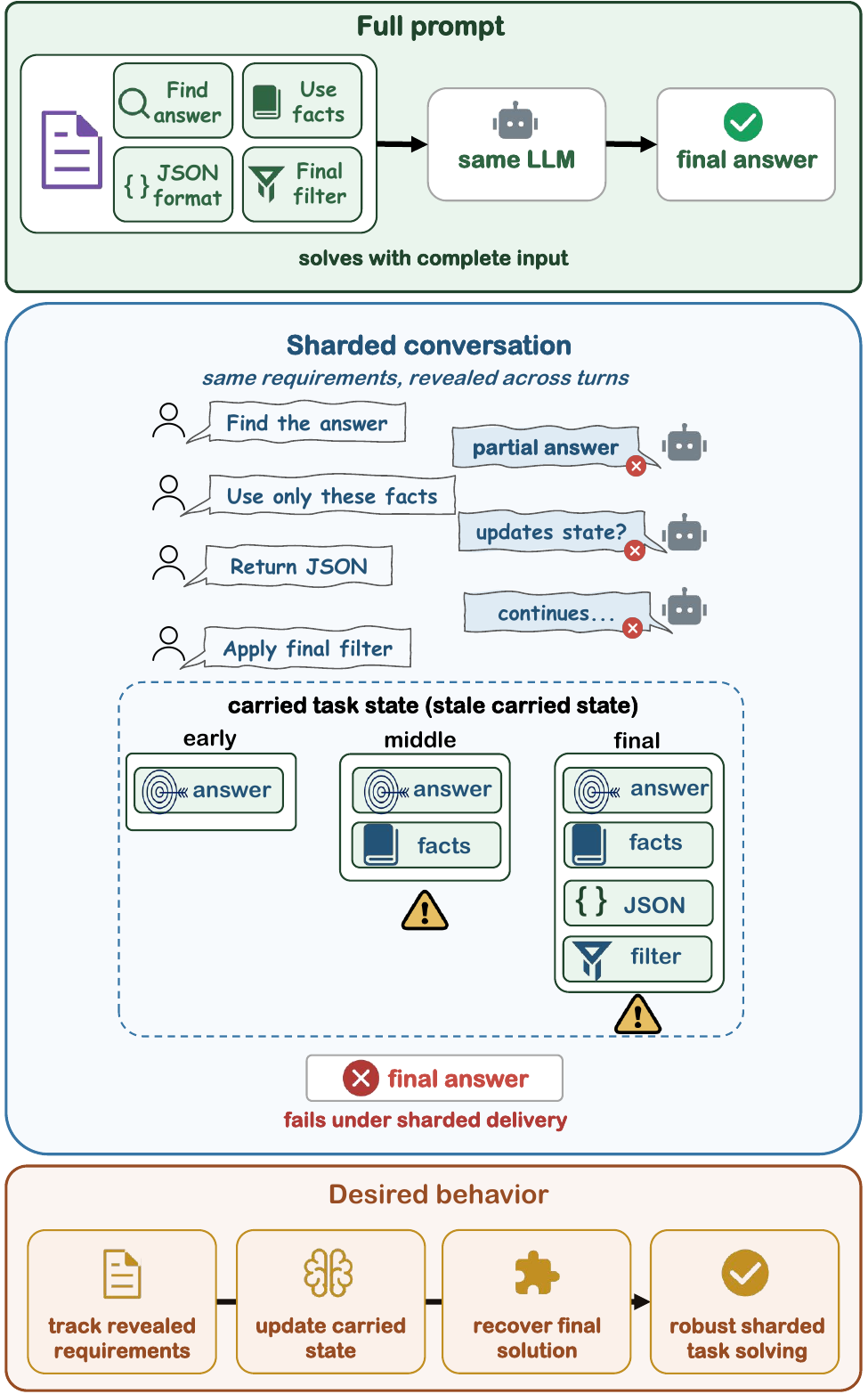}
    \caption{FULL-vs-SHARDED task delivery. The same requirements appear either
    in one complete prompt or across turns, where earlier assistant replies
    become part of the final context.}
    \label{fig:intro-problem}
\end{figure}

Large language models (LLMs) have shown strong ability to follow complex
instructions when all requirements are specified in a single
prompt~\citep{NEURIPS2022_b1efde53,
liu2024deepseek,qwen2025qwen25technicalreport,yang2025qwen3}. In real assistant
use, however, requirements are often accumulated, clarified, and revised across
multiple turns. This conversational setting exposes the
\textbf{Lost-in-Conversation (LiC) gap}: a model may solve a task in the FULL
view, where all requirements are provided at once, but fail in the SHARDED view,
where the same requirements are revealed progressively before the final
answer~\citep{laban2025llms}. Because the two views contain the same task
requirements, this gap is not simply a failure of single-turn competence, but a
failure to solve from a dialogue history shaped by both user turns and the
model's own earlier responses. This distinguishes LiC from ordinary
long-context access failures, where relevant information may be present but used
unevenly depending on where and how it appears~\citep{liu2024lost}, and aligns
with evidence that extended reasoning depends on intermediate planning and
optimization states rather than only on the final prompt~\citep{zheng2026pilot}.
Related benchmarks also show that language models can be sensitive to the order,
structure, and ambiguity of presented information~\citep{kang2026orderprobe,wang2025mucar}.

Existing LiC-oriented methods improve multi-turn robustness, but they do not
directly supervise the intermediate turns where a model can contaminate its own
future context. RLSTA~\citep{chen2026breaking}, for example, anchors
optimization to the single-turn full-prompt view through scalar GRPO feedback
and reports gains on contextual inertia. Other reward-based methods, including
RLAAR~\citep{li2025verifiable} and ICPO~\citep{wang2026icpo}, shape the policy
with verifiable-accuracy or illocution-calibrated rewards, while inference-time
methods such as Grounded Continuation~\citep{he2026grounded} introduce a runtime
verifier. These approaches provide answer-level rewards, full-view anchors, or
runtime checks, but they do not provide \textbf{dense supervision for
middle-turn assistant replies}. As a result, premature assumptions,
overconfident partial solutions, or stale intermediate states can still be
written into the dialogue history and later treated as context by the same
model. This creates a \textbf{compounding self-contamination effect}, analogous
to the distributional drift studied in sequential prediction and imitation
learning~\citep{ross2011reduction,bengio2015scheduled}.

We introduce \textbf{\method{}}, an on-policy self-distillation method designed
to reduce this self-contamination. The key idea is to train the model on
references that preserve the user-visible conversational prefix while removing
the assistant outputs that could otherwise bias the reference distribution. For
middle turns, \method{} constructs \textbf{history-cleaned references} by
deleting previous assistant replies but keeping the sharded user-side prefix
intact. This teaches the model how to respond to the current conversational
state without inheriting its own earlier mistakes. For answer turns, \method{}
distills from a paired full-view reference conditioned on the completed
user-side dialogue, aligning the final answer with the fully specified task. A
reliability weight further reduces the influence of middle-turn updates when
the sampled response diverges from the clean reference. The method requires no
verifier rewards, no additional state annotations, and no inference-time
scaffolding.\looseness=-1

We evaluate \method{} on Math, Actions, Database, and Code under the LiC
FULL-vs-SHARDED paired-view protocol with deterministic verifiers. On
\qwenSeven{}, \method{} improves average SHARDED accuracy from
\textbf{52.8 to 66.1} and raises the SHARDED/FULL ratio from
\textbf{66.5\% to 84.1\%}, while no FULL task changes by more than 2.3 points
relative to the base model. These results show that reducing self-contamination
can substantially narrow the LiC gap without compromising performance on fully
specified tasks.

Our contributions are threefold:
\begin{itemize}
    \item We identify self-contamination from intermediate assistant replies as
    a trainable contributor to the LiC gap, connecting multi-turn degradation to
    the compounding effects of self-conditioned dialogue histories.

    \item We propose \method{}, a turn-role-specific on-policy
    self-distillation method that uses history-cleaned middle-turn references,
    paired full-view answer references, and reliability-weighted supervision to
    reduce self-contamination without verifier rewards, state annotations, or
    inference-time scaffolding.

    \item We provide a controlled evaluation across math, tool-call generation,
    text-to-SQL, and Python code generation under the LiC paired-view protocol,
    showing substantial SHARDED gains while preserving FULL performance.
\end{itemize}

\section{Related Work}

\paragraph{Lost-in-Conversation and its mitigations.}
LiC introduced a controlled FULL/SHARDED comparison: a task that is solvable
from a complete single-turn instruction is split into user turns, and
performance is measured after the same information has been revealed
conversationally~\citep{laban2025llms}. The key implication is that
the multi-turn gap is not explained by lack of task competence alone. A model
that solves the FULL version can still fail in SHARDED form because of early
assumptions, premature answer attempts, over-reliance on previous attempts, and
weak recovery once the missing information arrives. A growing line of subsequent
studies characterizes this regime as a family of multi-turn state failures
rather than a single error type, including attention drift over goal tokens,
conversational inertia, and constraint or temporal-scope
violations~\citep{dongre2026attention,wan2026mitigating,
liu2026intent,canaverde2026sequor,kruthof2026models,
atri2026evaluating,zheng2026useless}. Existing mitigations mostly attack
this regime through scalar rewards, anchoring, or runtime checks:
verifiable-accuracy, abstention, or clarification
rewards~\citep{li2025verifiable,wang2026icpo,qiu2026pause}, single-turn-anchor
GRPO~\citep{shao2024deepseekmath,chen2026breaking}, and inference-time
dependency verification~\citep{he2026grounded}. \method{} is complementary but
adopts a different lens: it treats assistant-generated history itself as the
trainable failure mode and supplies dense, history-cleaned reference
distributions on the student's own SHARDED rollout.

\paragraph{On-policy distillation and context-conditioned self-teaching.}
On-policy distillation (OPD) trains on student-generated rollouts rather than
fixed reference outputs, reducing the train--test mismatch that can undermine
imitation in sequential decoding~\citep{agarwal2024policy}. Its self-distilled
variant (OPSD) uses the same model as student and teacher under different
contexts, with dense token-level matching on sampled
rollouts~\citep{zhao2026self}, while step-wise peer-based objectives provide
another form of intermediate trajectory signal~\citep{shi2026spader}. OPSD has
also been extended to context-conditioned
self-distillation~\citep{ye2026policy} and easier-context-to-harder-context
transfer for long-context generation~\citep{zhang2026opsdl}. Beyond text-only
generation, this paradigm has also supported short-to-long recovery of pruned
LLMs and regional-to-global transfer for fine-grained multimodal
perception~\citep{zhang2026shortopd,yuan2026visionopd}. Recent studies further show that on-policy
signals are not uniformly reliable: peer-rollout contexts, temporal curricula,
scope-aware schedules, and teacher-choice analyses all suggest that not every
on-policy update should be weighted
equally~\citep{yu2026multi,wang2026tcod,zheng2026scope,zhu2026many}. \method{} follows this
methodological line but adapts it to sharded conversations, where the main
failure mode is self-contamination from earlier assistant replies. Rather than
using a uniform reference context, it constructs turn-role-specific references:
history-cleaned supervision for middle turns and paired-view recovery
supervision for answer turns. A divergence-adaptive weight further modulates
middle-turn updates, connecting the method to reliability-aware OPD while
targeting the distinctive structure of multi-turn LiC failures.

Broader evaluations also probe robustness under cross-lingual factual
reasoning, multilingual hallucination mitigation, and multimodal cultural
grounding~\citep{weihua2026adamcot,weihua2025ccl,zheng2026mmac}.

\section{Method}
\label{sec:method}

\method{} is an on-policy self-distillation method for sharded multi-turn
tasks. The method trains on the student's own sharded rollouts, but computes
reference distributions under cleaner training-time contexts. This allows the
student to learn from the histories it actually induces, while preventing its
own earlier replies from contaminating the reference signal.

The training signal has three components. First, middle turns are supervised
with history-cleaned sharded references that preserve the user-visible prefix
but remove previous assistant replies. Second, the final answer turn is
supervised with a completed-task reference that preserves the full user-side
sharded dialogue and appends a canonical reserialization of the paired FULL
specification. Third, a separate FULL-view branch regularizes the model toward
its original full-prompt behavior. \method{} requires no verifier rewards, no
extra state labels, and no inference-time scaffolding.

\begin{figure*}[t]
    \centering
    \includegraphics[width=\textwidth]{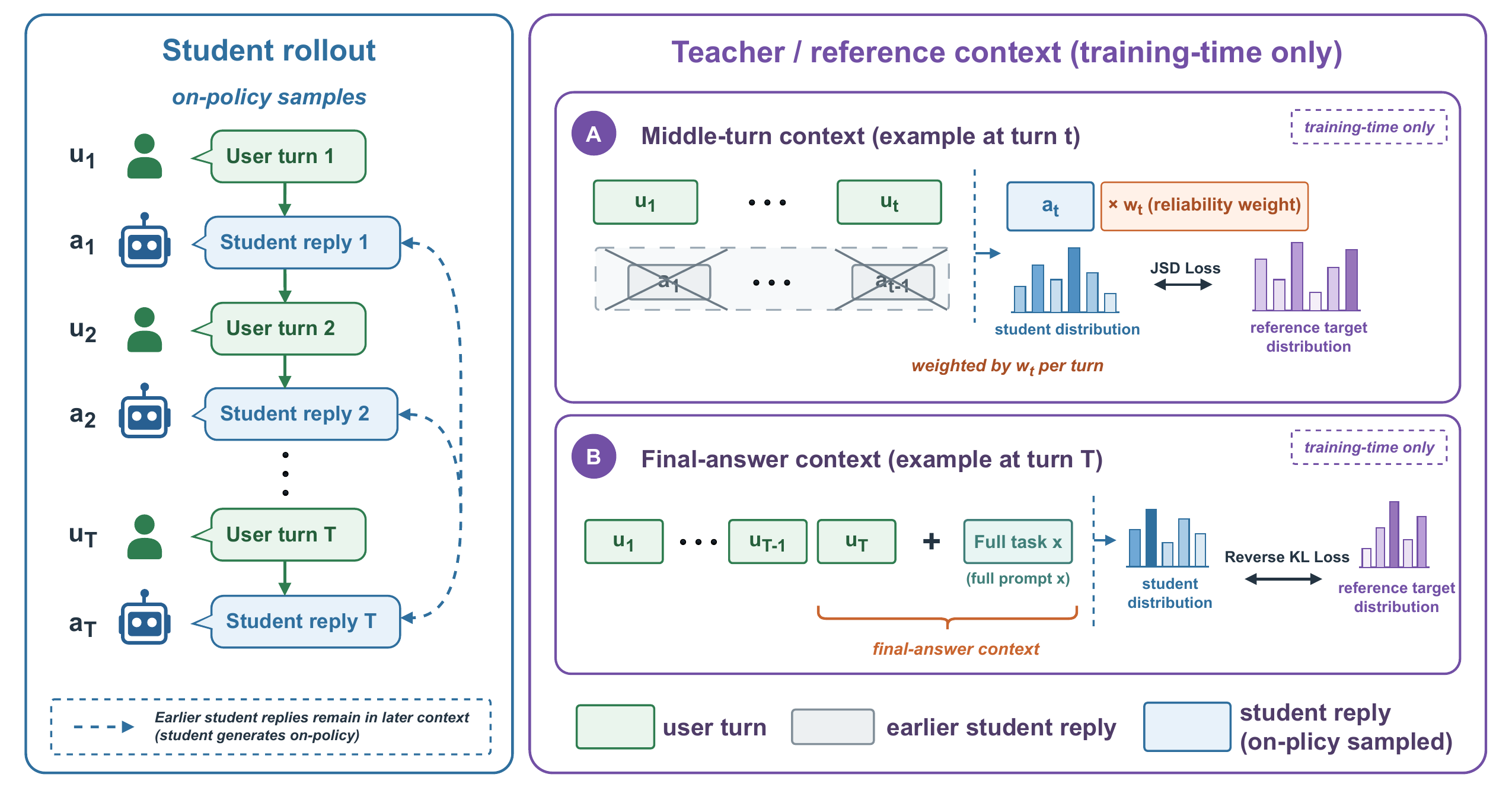}
    \caption{
    Overview of \method{}'s sharded training branch. Student samples are
    generated on-policy, while middle-turn and answer-turn losses use
    stop-gradient reference distributions computed from cleaner training-time
    contexts.
    }
    \label{fig:method-overview}
\end{figure*}

\subsection{Sharded On-Policy Setup}
\label{sec:method-setup}

Each training instance contains two paired views of the same task: a FULL
specification \(x\), and a sequence of user turns \(u_{1:T}\) that reveal the
same requirements incrementally, following the FULL/SHARDED setting of
Lost-in-Conversation~\citep{laban2025llms}. The final user turn \(u_T\) is the
answer turn. Assistant replies before \(T\) are not final outputs, but they
remain in the dialogue history and therefore shape the context for later turns.

The student policy \(\pi_\theta\) is rolled out on the sharded dialogue
on-policy. At turn \(t\), the student context is
\begin{equation}
    C_t^S = (u_1,\hat a_1,\ldots,u_{t-1},\hat a_{t-1},u_t),
\end{equation}
and the student samples
\begin{equation}
    \hat a_t \sim \pi_\theta(\cdot \mid C_t^S),
    \qquad t=1,\ldots,T .
\end{equation}
Thus training is performed on the same self-conditioned history distribution
that can produce SHARDED failures, rather than on a clean offline transcript.

For a sampled assistant response
\(\hat a=(\hat y_1,\ldots,\hat y_n)\) generated under student context \(C^S\),
all distillation losses use the same token-level form:
\begin{equation}
    \mathcal{D}_{\beta}(C^S,C^{\mathrm{ref}};\hat a)
    =
    \frac{1}{n}\sum_{i=1}^{n}
    d_{\beta}\!\left(P_i^{\mathrm{ref}}, P_i^S\right),
\label{eq:turn-distill}
\end{equation}
where
\begin{equation}
    P_i^S =
    \pi_\theta(\cdot \mid C^S,\hat y_{<i})
\end{equation}
is the student next-token distribution, and
\begin{equation}
    P_i^{\mathrm{ref}} =
    \operatorname{sg}\!\left[
    \pi_{\theta^-}(\cdot \mid C^{\mathrm{ref}},\hat y_{<i})
    \right]
\end{equation}
is the stop-gradient reference distribution evaluated under a role-specific
reference context \(C^{\mathrm{ref}}\). Here \(\pi_{\theta^-}\) is an
exponential-moving-average (EMA) reference policy~\citep{agarwal2024policy,zhao2026self}.

The divergence \(d_\beta\) follows the generalized Jensen--Shannon divergence
used in OPSD. For \(0<\beta<1\), let
\begin{equation}
    m_\beta = \beta p + (1-\beta)q ,
\end{equation}
and define
\begin{equation}
\begin{aligned}
    d_\beta(p,q)
    &=
    \beta\,\mathrm{KL}(p\,\|\,m_\beta) \\
    &\quad +(1-\beta)\,\mathrm{KL}(q\,\|\,m_\beta).
\end{aligned}
\label{eq:dbeta}
\end{equation}
For the endpoints, we follow the OPSD convention and set
\(d_0(p,q)=\mathrm{KL}(p\|q)\) and
\(d_1(p,q)=\mathrm{KL}(q\|p)\). In Eq.~\ref{eq:turn-distill},
\(p=P_i^{\mathrm{ref}}\) and \(q=P_i^S\). In our reported configuration,
\(\beta_{\mathrm{mid}}=0.5\) and
\(\beta_{\mathrm{final}}=\beta_{\mathrm{full}}=1.0\), so the middle branch uses
the symmetric Jensen--Shannon loss, while the answer and FULL branches use
reverse KL. Gradients pass only through the student distribution \(P_i^S\).

The remainder of this section specifies the reference context
\(C^{\mathrm{ref}}\) for each turn role.

\subsection{History-Cleaned Middle-Turn Supervision}
\label{sec:method-middle}
\label{sec:method-primitive}
\label{sec:method-middle-reference}
\label{sec:method-adaptive}

Middle turns are partial-information states. Supervising them with a full-task
reference can encourage premature solution tokens, while supervising them with
a reference that conditions on the same assistant-generated history can inherit
the student's earlier mistakes. \method{} therefore uses a history-cleaned
reference for every eligible middle turn:
\begin{equation}
    C_{t,\mathrm{mid}}^{\mathrm{ref}}
    =
    (\sigma_{\mathrm{sys}}, u_1, u_2, \ldots, u_t),
    \qquad t<T,
\end{equation}
where \(\sigma_{\mathrm{sys}}\) denotes the task-specific system block shared
by the FULL and SHARDED views; no previous assistant replies, unrevealed user
turns, or full-task content appear in the reference. Intuitively, the
reference asks what the EMA reference policy assigns to the student's sampled
middle response when the model's own earlier outputs are not part of the
conditioning context.

The local middle-turn loss is
\begin{equation}
    \ell_t^{\mathrm{mid}}
    =
    \mathcal{D}_{\beta_{\mathrm{mid}}}
    (C_t^S,C_{t,\mathrm{mid}}^{\mathrm{ref}};\hat a_t).
\label{eq:middle-local-loss}
\end{equation}
The sampled tokens are generated from the student's actual sharded context
\(C_t^S\), keeping the update on-policy. In contrast, the reference distribution
is evaluated under the history-cleaned context
\(C_{t,\mathrm{mid}}^{\mathrm{ref}}\), where previous assistant replies have
been removed. Thus, the loss aligns the student's middle-turn behavior with
what the EMA reference policy would assign under a clean user-side prefix,
rather than under a context already shaped by the student's earlier outputs.
This encourages middle-turn replies that are less likely to introduce
self-generated errors into later dialogue context.

\paragraph{Reliability weighting.}
Not every sampled middle turn provides an equally reliable reference signal:
on-policy rollouts can contain low-quality questions, verbose partial answers,
or histories that the clean sharded prefix would not support. We therefore
weight each middle turn by the realized student--reference log-probability
discrepancy on the sampled tokens. Let \(\mathcal{I}\) collect the pre-answer
turns whose sampled replies are non-empty and retained for token-level loss.
If \(\mathcal{I}\) is empty, the sharded branch skips the middle loss for
that example. For each \(t\in\mathcal{I}\), define a detached scalar
\begin{equation}
    \delta_t
    =
    \operatorname{sg}\!\left[
        \tfrac{1}{m_t}\!\sum_{i=1}^{m_t}\!
        \big|
        \log P_{t,i}^{S}(\hat y_{t,i})
        -\log P_{t,i}^{\mathrm{ref}}(\hat y_{t,i})
        \big|
    \right],
\label{eq:middle-divergence}
\end{equation}
where \(P_{t,i}^{S}\) and \(P_{t,i}^{\mathrm{ref}}\) are the student and
reference next-token distributions at position \(i\) of middle turn \(t\), and
\(m_t\) is the number of supervised tokens in the sampled middle reply. Intuitively,
\(\delta_t\) tracks how far the realized middle response has drifted from a
clean-history view of the same on-policy state. Let
\(\mathcal{I}_{\le t}=\{j\in\mathcal{I}:j\le t\}\) and
\(\delta_j^+=\max(\delta_j,\epsilon)\), where \(\epsilon=10^{-6}\) provides
numerical stability. We turn
\(\delta_t\) into a bounded weight using the median discrepancy over eligible
turns up to \(t\),
\begin{equation}
\begin{aligned}
    \tilde{\delta}_t &= \operatorname{median}_{j\in\mathcal{I}_{\le t}}(\delta_j^+), \\
    w_t &= \operatorname{sg}\!\left[
        \frac{\tilde{\delta}_t}{\tilde{\delta}_t+\delta_t^+}
    \right] \in (0,1] ,
\end{aligned}
\label{eq:middle-weight}
\end{equation}
The form gives \(w_t = 1/2\)
at the median and decays approximately as \(1/\delta_t\) for outliers, so middle supervision
is strongest when the sampled response stays close to the history-cleaned view
and more conservative when the trajectory has drifted far from it. This follows
the broader OPSD observation that on-policy reference signals vary in quality
across samples~\citep{wang2026tcod,zheng2026scope,zhu2026many}.

The middle-turn objective for a realized trajectory is the mean over eligible
turns,
\begin{equation}
    \bar{\mathcal{L}}_{\mathrm{mid}}
    =
    \frac{1}{|\mathcal{I}|}
    \sum_{t\in\mathcal{I}}
    \alpha_t w_t\ell_t^{\mathrm{mid}} ,
\label{eq:middle-mean-loss}
\end{equation}
where \(\alpha_t\) is a middle-turn scale, set to \(1.0\) for all eligible
turns in our experiments. For efficiency, each trajectory samples one
eligible middle turn,
\(\tau \sim \operatorname{Unif}(\mathcal{I})\), and uses
\begin{equation}
    \mathcal{L}_{\mathrm{mid}}
    =
    \alpha_{\tau} w_{\tau}\ell_{\tau}^{\mathrm{mid}} ,
\label{eq:middle-sampled-loss}
\end{equation}
which is an unbiased single-turn estimator of Eq.~\ref{eq:middle-mean-loss}
under uniform \(\tau\) when the per-turn weights are treated as detached
scalars.

\subsection{Answer-Turn Recovery and Full-Prompt Preservation}
\label{sec:method-recovery}
\label{sec:method-final}
\label{sec:method-full}

Middle supervision alone is not enough: the target is task completion after
the required information arrives. At turn \(T\), the student's context
\(C_T^S\) may still contain earlier replies that are incomplete, speculative,
or inconsistent with later user information. A reference that conditions on
the same assistant-generated history would inherit those errors, while a
reference that conditions only on the full task view would ignore the sharded
delivery the student must handle. \method{} therefore builds a training-only
answer-turn reference context
\begin{equation}
    C_{T,\mathrm{ans}}^{\mathrm{ref}} =
    (\sigma_{\mathrm{sys}}, u_1, \ldots, u_T, \sigma_{\mathrm{canon}}(x)),
\end{equation}
which keeps the completed user-side sharded dialogue and appends a
deterministic canonical reserialization \(\sigma_{\mathrm{canon}}(x)\) of the
paired FULL specification \(x\) as an additional message, with previous
student replies removed before serialization (full role-assignment details
in App.~\ref{sec:appendix-reference-views}). The canonical block is a
reserialization of requirements already present in the paired task; it does
not add a gold answer or verifier signal. The answer-turn loss is
\begin{equation}
    \mathcal{L}_{\mathrm{final}}
    =
    \mathcal{D}_{\beta_{\mathrm{final}}}
    (C_T^S,C_{T,\mathrm{ans}}^{\mathrm{ref}};\hat a_T) ,
\label{eq:final-loss}
\end{equation}
and teaches solution recovery at the natural answer turn: the reference gives
the EMA reference policy a cleaner organization of the same completed task,
while the student loss is still applied to tokens sampled from the ordinary
sharded dialogue.

\paragraph{FULL-view preservation.}
Because the sharded branch reshapes behavior under sharded delivery, we
preserve FULL-view capability with an EMA anchor on the FULL view. Let
\(C^F\) be the full task context built from \(x\) and sample
\(\hat a^F\sim\pi_\theta(\cdot\mid C^F)\). The student and reference share the
same FULL context here, so
\begin{equation}
    \mathcal{L}_{\mathrm{full}}
    =
    \mathcal{D}_{\beta_{\mathrm{full}}}(C^F,C^F;\hat a^F)
\label{eq:full-loss}
\end{equation}
is an EMA self-anchor rather than a source of new task information; it
discourages the model from trading away competence it already exhibits when
all requirements are provided at once.

\subsection{Training Objective}
\label{sec:method-training}

The training objective has two branches. The sharded trajectory branch
combines answer-turn recovery with weighted history-cleaned middle-turn
distillation, and the FULL-view branch acts as a preservation regularizer:
\begin{equation}
\begin{aligned}
    \mathcal{L}_{\mathrm{train}} &=
    (1\!-\!z)(\mathcal{L}_{\mathrm{final}}+\mathcal{L}_{\mathrm{mid}})
    + z\,\mathcal{L}_{\mathrm{full}}, \\
    z &\sim \operatorname{Bernoulli}(\rho),
\end{aligned}
\label{eq:maigo-objective}
\end{equation}
where \(\rho\) controls how often the preservation branch is sampled; with
probability \(1-\rho\), the update uses the sharded trajectory branch. Taking
the expectation over \(z\) recovers the mixture of the two
branches. After each optimizer step, the EMA reference is updated as
\(\theta^- \leftarrow \lambda\theta^- + (1-\lambda)\theta\) with decay
\(\lambda\). Algorithm~\ref{alg:maigo-training} summarizes a training step.

\begin{algorithm}[t]
\caption{One \method{} training step}
\label{alg:maigo-training}
\footnotesize
\setlength{\tabcolsep}{0.35em}
\renewcommand{\arraystretch}{1.08}
\begin{tabular}{@{}r>{\raggedright\arraybackslash}p{0.78\linewidth}@{}}
\toprule
\multicolumn{2}{@{}p{0.94\linewidth}@{}}{\textbf{Input:} paired views
\((x,u_{1:T})\), student \(\pi_\theta\), EMA reference policy
\(\pi_{\theta^-}\), FULL-branch rate \(\rho\), EMA decay \(\lambda\).}\\
\multicolumn{2}{@{}p{0.94\linewidth}@{}}{\textbf{Output:} updated student
parameters \(\theta\).}\\
\midrule
1 & Sample \(z\sim\operatorname{Bernoulli}(\rho)\).\\
2 & \textbf{if} \(z=1\) \textbf{then}\\
3 & \quad Sample \(\hat a^F\sim\pi_\theta(\cdot\mid C^F)\) and optimize
\(\mathcal{L}_{\mathrm{full}}\).\\
4 & \textbf{else}\\
5 & \quad Roll out
\(\hat a_t\sim\pi_\theta(\cdot\mid C_t^S)\), \(t=1,\ldots,T\).\\
6 & \quad For eligible middle turns, build
\(C_{t,\mathrm{mid}}^{\mathrm{ref}}\) and compute
\(\ell_t^{\mathrm{mid}}\), \(\delta_t\), \(w_t\).\\
7 & \quad If \(\mathcal{I}\!\neq\!\emptyset\), sample
\(\tau\!\in\!\mathcal{I}\) and set
\(\mathcal{L}_{\mathrm{mid}}\!=\!\alpha_\tau w_\tau\ell_\tau^{\mathrm{mid}}\);
else set \(\mathcal{L}_{\mathrm{mid}}=0\).\\
8 & \quad Build \(C_{T,\mathrm{ans}}^{\mathrm{ref}}\), compute
\(\mathcal{L}_{\mathrm{final}}\), and optimize
\(\mathcal{L}_{\mathrm{final}}+\mathcal{L}_{\mathrm{mid}}\).\\
9 & \textbf{end if}\\
10 & Update \(\theta^- \leftarrow \lambda\theta^-+(1-\lambda)\theta\).\\
\bottomrule
\end{tabular}
\end{algorithm}

\section{Experiments}
\label{sec:experiments}

\begin{table*}[!t]
\centering
\footnotesize
\setlength{\tabcolsep}{3pt}
\renewcommand{\arraystretch}{1.05}
\begin{tabular}{@{}ll ccccc|ccccc|c@{}}
\toprule
& & \multicolumn{5}{c|}{FULL accuracy (\%)}
& \multicolumn{5}{c|}{SHARDED accuracy (\%)}
& \\
\cmidrule(lr){3-7}\cmidrule(lr){8-12}
Model & Method
& Math & Actions & Database & Code & Avg.
& Math & Actions & Database & Code & Avg.
& S/F (\%) \\
\midrule
\multirow{5}{*}{\qwenSeven{}}
 & Base    & 89.1 & 82.2 & 61.2 & 84.8 & 79.3 & 70.5 & 17.8 & 42.2 & 80.6 & 52.8 & 66.5 \\
 & SFT     & 89.0 & 79.3 & 61.4 & 80.5 & 77.5 & 72.1 & 18.2 & 44.1 & 77.7 & 53.0 & 68.4 \\
 & GRPO    & 89.8 & 81.6 & 59.4 & 85.1 & 79.0 & 73.2 & 21.5 & 42.5 & 83.5 & 55.2 & 69.9 \\
 & RLSTA   & 90.1 & 81.8 & 59.8 & 86.0 & 79.4 & 77.0 & 24.5 & 45.0 & 84.0 & 57.6 & 72.5 \\
 & \method{}
          & 90.3 & 82.1 & 59.6 & 82.5 & 78.6
          & \textbf{82.5} & \textbf{46.8} & \textbf{50.0} & \textbf{85.2}
          & \textbf{66.1} & \textbf{84.1} \\
\midrule
\multirow{5}{*}{\qwenThree{}}
 & Base    & 81.4 & 72.8 & 50.7 & 59.0 & 66.0 & 59.9 & 12.0 & 32.9 & 69.4 & 43.5 & 66.0 \\
 & SFT     & 80.2 & 68.5 & 50.8 & 53.5 & 63.3 & 56.5 & 12.2 & 36.6 & 67.8 & 43.3 & 68.4 \\
 & GRPO    & 81.6 & 70.9 & 51.7 & 51.1 & 63.8 & 61.5 & 15.8 & 34.0 & 73.3 & 46.1 & 72.3 \\
 & RLSTA   & 82.3 & 73.2 & 52.0 & 55.0 & 65.6 & 64.0 & 17.3 & 37.0 & 73.5 & 47.9 & 73.1 \\
 & \method{}
          & 81.1 & 77.5 & 51.9 & 59.6 & 67.5
          & \textbf{72.0} & \textbf{40.0} & \textbf{44.9} & \textbf{76.1}
          & \textbf{58.2} & \textbf{86.3} \\
\bottomrule
\end{tabular}
\caption{Main results on GSM8K~\citep{cobbe2021training}
(Math; $N{=}1319$), BFCL~\citep{patil2025berkeley} parallel/parallel-multiple
(Actions; $N{=}382$), Spider~\citep{yu2018spider} no-easy (Database;
$N{=}715$), and HumanEval~\citep{chen2021evaluating} (Code; $N{=}164$).
Values are mean accuracy (\%). Fine-tuned rows average four training seeds;
Base averages four evaluation seeds. \emph{Avg.} is the unweighted task mean;
\emph{S/F} is the SHARDED/FULL ratio of \emph{Avg.}.}
\label{tab:main-results}
\end{table*}

We report SHARDED accuracy as the primary metric and FULL accuracy as a
complete-prompt preservation check.

\subsection{Evaluation Setup}
\label{sec:exp-protocol}

We follow the Lost-in-Conversation FULL-vs-SHARDED paired-view protocol:
each item has a complete-prompt view and a sharded conversational view with
the same requirements revealed across user turns~\citep{laban2025llms}.
Following LiC and subsequent LiC-based training studies, SHARDED evaluation
uses a model-assisted user simulator that reveals at most one unrevealed shard
per turn, lets the evaluated assistant respond, and then decides whether an
answer attempt should be scored~\citep{laban2025llms,li2025verifiable}.
All methods use the same fixed \deepseekvthree{} simulator for user turns and
answer-attempt classification~\citep{liu2024deepseek}, along with the same
shard files, prompts, seeds, and deterministic scorers.
Fine-tuning uses disjoint
sharded training sets of 6{,}400 examples for Math, Actions, and Code, and
6{,}397 examples for Database. The Code training pool combines filtered
DeepMind Code Contests examples with deduplicated APPS training examples,
while evaluation uses the full 164-example HumanEval split; Appendix~\ref{sec:appendix-tasks}
gives the exact source counts and construction details. Outputs that fail to parse, are empty, violate the
response protocol, or are truncated by the generation budget count as
incorrect. For \method{}, these scorers are evaluation-only, not training
rewards.

\paragraph{Tasks and metrics.}
We use the four objectively scored task families from LiC---the same subset adopted
by recent LiC-based training
studies~\citep{li2025verifiable,chen2026breaking,qiu2026pause}---and exclude the
two judge-scored ones (summary, captioning) so that every cell is auditable
by a deterministic verifier. \textbf{Math} is instantiated by LiC-style GSM8K
with numeric matching~\citep{cobbe2021training}; \textbf{Actions} by BFCL
parallel/parallel-multiple tool calls with deterministic AST
matching~\citep{patil2025berkeley}; \textbf{Database} by fixed-schema sharded
text-to-SQL on Spider no-easy with execution matching~\citep{yu2018spider};
and \textbf{Code} by HumanEval Python function generation scored by unit
tests~\citep{chen2021evaluating}. Together, they cover numeric, tool-call,
SQL, and program outputs, giving deterministic probes of different
contamination channels.

\paragraph{Models and baselines.}
The primary comparison uses \qwenSeven{}, with 3B rows as a
smaller-backbone validation~\citep{qwen2025qwen25technicalreport}. Baselines are per-task SFT,
GRPO with sparse verifiable rewards~\citep{shao2024deepseekmath}, and RLSTA, which
uses the stronger single-turn view as a GRPO anchor~\citep{chen2026breaking}. All
trained methods use Low-Rank Adaptation (LoRA) adapters~\citep{hu2022lora};
adapter weights are
merged into the base model before evaluation. All methods use the same paired
FULL/SHARDED prompts under a matched 100-step fine-tuning budget. We report
the mean of four training seeds for each fine-tuned method.
Appendix~\ref{sec:appendix-optimization} reports
optimizer, context, and generation settings.

\subsection{Main Results}
\label{sec:exp-main}

Table~\ref{tab:main-results} reports per-task FULL and SHARDED accuracy, their
averages, and the SHARDED-to-FULL ratio. The base model is strong on FULL but
much weaker once the same requirements arrive through turns;
on \qwenSeven{} Actions, accuracy drops from 82.2 FULL to 17.8 SHARDED, and the
base S/F ratio is 66.5\%, close to the roughly 60\% ratio reported by LiC for
closed-source models~\citep{laban2025llms}. \method{} improves every 7B
SHARDED task, with the largest gain on Actions (+29.0). Averaged over the four
7B task families, SHARDED accuracy rises from 52.8 to 66.1 and the S/F ratio
rises from 66.5\% to 84.1\%, while average FULL changes only from 79.3 to
78.6, with no individual task FULL changing by more than 2.3 points.

\paragraph{One-adapter mixed-task feasibility.}
The headline results use one LoRA adapter per task to provide a controlled
comparison of training signals. To test whether \method{} also supports a
single general-purpose adapter, we trained a balanced mixture of 25,600
examples (6,400 per task) for one complete pass. The resulting adapter reaches
68.33\% SHARDED and 78.29\% FULL macro accuracy, with nontrivial SHARDED
performance on all four task families (Table~\ref{tab:mixed-feasibility}).
This is a feasibility check rather than an equal-compute comparison with the
per-task adapters: the mixed run uses a global batch size of 256, whereas the
headline runs use 64.

\begin{table}[t]
\centering
\small
\setlength{\tabcolsep}{5pt}
\renewcommand{\arraystretch}{1.05}
\begin{tabular}{@{}lrr@{}}
\toprule
Task & FULL & SHARDED \\
\midrule
Actions   & 81.41 & 50.52 \\
GSM8K     & 88.55 & 84.23 \\
Database  & 60.28 & 54.41 \\
HumanEval & 82.93 & 84.15 \\
\midrule
Macro     & 78.29 & 68.33 \\
\bottomrule
\end{tabular}
\caption{One-pass mixed-task \method{} adapter trained on a balanced 25,600-example
mixture. Values are accuracy (\%).}
\label{tab:mixed-feasibility}
\end{table}

\paragraph{Direct history intervention.}
To isolate the proposed self-contamination mechanism, we replayed each Base
model terminal trace with the same user turns and decoder, once with its
naturally accumulated assistant history and once after deleting all prior
assistant replies. The clean-history intervention substantially improves
accuracy on three task families and is directional on HumanEval
(Table~\ref{tab:history-intervention}); exact paired McNemar tests give
\(p=7.88\times10^{-40}\), \(0.00219\), and
\(7.03\times10^{-34}\) for Actions, Database, and GSM8K, respectively.
Because the intervention changes only the recorded assistant history, these
results provide direct evidence that self-generated history is a contributor
to the LiC gap, while not implying that it explains every failure mode.

\begin{table}[t]
\centering
\small
\setlength{\tabcolsep}{3.5pt}
\renewcommand{\arraystretch}{1.05}
\begin{tabular}{@{}lrrr@{}}
\toprule
Task & Contaminated & Clean history & Gain \\
\midrule
Actions   & 16.75 & 54.45 & +37.70 \\
Database  & 38.18 & 43.22 & +5.03 \\
GSM8K     & 60.20 & 78.01 & +17.82 \\
HumanEval & 74.39 & 79.88 & +5.49 \\
\bottomrule
\end{tabular}
\caption{Fixed-Base history intervention on identical terminal traces. Values
are accuracy (\%); all outputs are scored directly with deterministic
verifiers, without the answer-attempt classifier or extractor.}
\label{tab:history-intervention}
\end{table}

\paragraph{Direct-verifier rescoring.}
We further test whether the observed improvement could be attributed to the
model-assisted answer-attempt classifier or answer extractor. On a fixed set
of GSM8K terminal outputs, we bypass both components and score the same
responses directly with deterministic whole-response matching. Under this
setting, \method{} reaches 79.91\% accuracy, compared with 63.00\% for the Base
model, yielding a paired gain of 16.91 points ($p=7.9\times10^{-32}$). The
corresponding gap under the standard evaluation pipeline is 12.21 points.
Thus, removing the model-assisted scoring components increases rather than
attenuates the relative gain, supporting the interpretation that \method{}
improves task completion rather than merely adherence to the expected answer
surface.

\paragraph{Comparison to supervised and RL baselines.}
The baseline rows show limits of standard fine-tuning signals in
this setting. On \qwenSeven{}, SFT changes average SHARDED accuracy
marginally. GRPO and RLSTA raise the 7B SHARDED average to 55.2 and 57.6, but
they leave a large residual gap on Actions, where early tool arguments and
partial calls remain in later context.
\method{} instead supplies dense token-level
supervision on sampled sharded rollouts under cleaned reference histories,
which better targets assistant replies that will become part of the future
state.

\paragraph{Task-level behavior.}
Actions benefits most because tool calls and shared arguments accumulate across
turns. Math and Database show smaller but clear improvements when facts or SQL
constraints arrive conversationally. Code is the one family where SHARDED can
exceed FULL for some rows. This is possible because FULL is not an absolute
upper bound: the HumanEval SHARDED view reveals the same function requirements
as short conversational constraints, which can make the required signature and
late conditions explicit near generation time before the same unit-test scorer
is applied. The relevant pattern is therefore SHARDED improvement together
with FULL preservation; by that measure, \method{} gives stable Code gains
(+4.6 on 7B and +6.7 on 3B) without degrading average FULL behavior.

Across tasks, gains are largest when early replies commit to partial
interpretations that later turns must revise, especially for accumulating tool
arguments. Math, Database, and Code show task-specific recovery patterns,
motivating MAIGO's role-specific middle-turn and answer-turn references.

\begin{table*}[!t]
\centering
\footnotesize
\setlength{\tabcolsep}{3pt}
\renewcommand{\arraystretch}{1.08}
\begin{tabular}{@{}l*{4}{cc}|cccc@{}}
\toprule
& \multicolumn{2}{c}{Math}
& \multicolumn{2}{c}{Actions}
& \multicolumn{2}{c}{Database}
& \multicolumn{2}{c|}{Code}
& \multicolumn{4}{c}{Average} \\
\cmidrule(lr){2-3}\cmidrule(lr){4-5}\cmidrule(lr){6-7}\cmidrule(lr){8-9}\cmidrule(l){10-13}
Training signal
& F & S & F & S & F & S & F & S & F & S & \(\Delta\)S & S/F (\%) \\
\midrule
\method{} (default)
& \textbf{90.3} & \textbf{82.5} & 82.1 & 46.8 & 59.6 & \textbf{50.0} & 82.5 & \textbf{85.2}
& 78.6 & \textbf{66.1} & -- & 84.1 \\
\midrule
$-$ middle loss
& 89.4 & 76.1 & 83.0 & 41.5 & 60.4 & 47.8 & 84.2 & 81.0
& 79.2 & 61.6 & -4.5 & 77.7 \\
$-$ adaptive weighting (uniform)
& 89.1 & 81.7 & 81.4 & 44.6 & \textbf{61.3} & 45.6 & \textbf{85.5} & 79.3
& \textbf{79.3} & 62.8 & -3.4 & 79.1 \\
$-$ FULL-only answer ref.
& 88.6 & 82.0 & 82.3 & \textbf{48.1} & 57.2 & 49.1 & 82.3 & 84.6
& 77.6 & 65.9 & -0.2 & \textbf{85.0} \\
$-$ preservation branch
& 89.2 & 78.7 & \textbf{83.2} & 48.0 & 56.9 & 47.4 & 82.0 & 81.1
& 77.8 & 63.8 & -2.3 & 82.0 \\
\bottomrule
\end{tabular}
\caption{Component ablations on \qwenSeven{}. F and S are FULL and SHARDED
accuracy (\%); \(\Delta\)S is the change in average SHARDED accuracy relative
to \method{}, and S/F is the ratio of average S to average F. Each ablation
changes one component and keeps the other settings fixed; values are four-seed
means.}
\label{tab:component-ablations}
\end{table*}

\begin{figure}[t]
    \centering
    \includegraphics[width=\columnwidth]{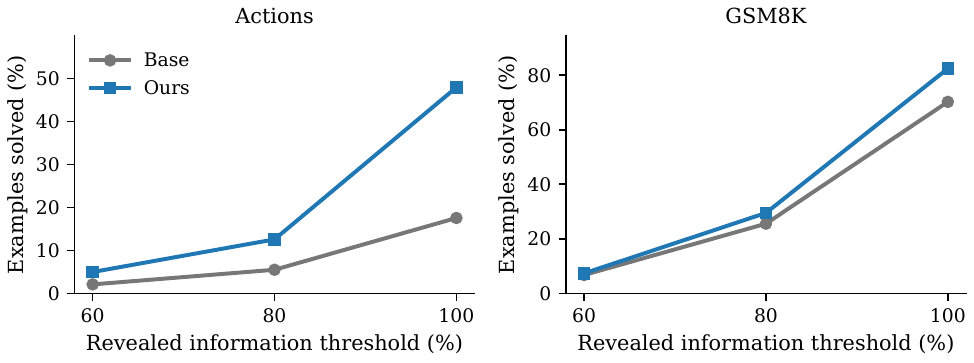}
    \caption{Single-seed reveal-time diagnostic on \qwenSeven{} SHARDED traces;
    the 100\% point equals final SHARDED accuracy.}
    \label{fig:first-correct-reveal}
\end{figure}

\paragraph{Trajectory diagnostic.}
Figure~\ref{fig:first-correct-reveal} asks when each SHARDED example first
becomes correct during the rollout. On Actions, the
full-reveal gain is much larger than the gain at 80\% reveal (\(+30.4\) vs.
\(+7.1\) points); Math shows a smaller but consistent shift. This indicates
that much of the improvement appears after all requirements have arrived, when
the student must answer from its accumulated history. The main FULL results
also rule out a simple tradeoff against single-turn performance.

\paragraph{Smaller model check.}
The \qwenThree{} rows replicate the pattern: average SHARDED accuracy rises
from 43.5 to 58.2 and S/F from 66.0\% to 86.3\%, while average FULL stays close
to the base model; ablations remain on the 7B suite.

\begin{table}[t]
\centering
\footnotesize
\setlength{\tabcolsep}{5pt}
\renewcommand{\arraystretch}{1.08}
\begin{tabular}{@{}lccc@{}}
\toprule
Task & Base F/S & \method{} F/S & \(\Delta\)S \\
\midrule
Math & 83.7 / 61.3 & 83.9 / 79.2 & +17.9 \\
Actions & 78.9 / 29.2 & 80.2 / 54.3 & +25.1 \\
Database & 56.6 / 37.2 & 58.2 / 50.8 & +13.6 \\
Code & 65.6 / 75.0 & 72.1 / 80.2 & +5.2 \\
\midrule
Avg. & 71.2 / 50.7 & 73.6 / 66.1 & +15.5 \\
\bottomrule
\end{tabular}
\caption{Cross-backbone generality on Llama-3.1-8B-Instruct. F/S is
FULL/SHARDED accuracy (\%); values are four-seed means.}
\label{tab:llama-check}
\end{table}

\paragraph{Cross-backbone generality.}
Table~\ref{tab:llama-check} gives a Base-vs-\method{} comparison on
Llama-3.1-8B-Instruct~\citep{grattafiori2024llama}. SHARDED accuracy improves on four task families,
raising the average from 50.7 to 66.1 and S/F from 71.2\% to 89.8\%, while
average FULL accuracy also increases. This supports a history-cleaned
self-distillation effect beyond the Qwen backbones used for the
matched-baseline study.

\subsection{Component Ablations}
\label{sec:exp-ablations}

Table~\ref{tab:component-ablations} isolates the main components. Removing the
middle-turn loss causes the largest degradation: average SHARDED
accuracy falls by 4.5 points and the S/F ratio drops from 84.1\% to 77.7\%,
showing that final-answer supervision alone misses useful signal from
history-cleaned intermediate replies. Flattening the adaptive weight to a
uniform value gives a similar 3.4-point SHARDED drop; although this variant has
a slightly higher FULL average, it does not recover the same sharded behavior.
Thus the trajectory-local weight has an effect beyond simply scaling the
middle loss. A FULL-only answer reference is the strongest ablation in average
SHARDED accuracy (65.9 vs.\ 66.1), slightly exceeding \method{} on Actions but
lowering mean FULL (77.6 vs.\ 78.6). The preservation branch acts as a
regularizer: removing it lowers both averages, with Actions the only task
where removal improves SHARDED accuracy.

\subsection{Hyperparameter Sensitivity}
\label{sec:exp-rho-sensitivity}

\begin{figure}[t]
    \centering
    \includegraphics[width=\columnwidth]{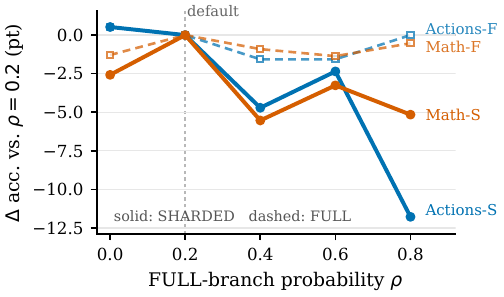}
    \caption{Single-seed sensitivity to the FULL-preservation probability
    \(\rho\) on \qwenSeven{}; values are accuracy changes from the default
    \(\rho=0.2\).}
    \label{fig:rho-sensitivity}
\end{figure}

The preservation branch is controlled by the probability \(\rho\) of sampling a
FULL-view training step. Figure~\ref{fig:rho-sensitivity} sweeps
\(\rho\in\{0,0.2,0.4,0.6,0.8\}\) on Actions and Math. The default
\(\rho=0.2\) is already near the best SHARDED result;
larger values do not improve the conversational setting and can reduce SHARDED
accuracy, especially on Actions. FULL accuracy is comparatively stable, so the
main effect of large \(\rho\) is weaker SHARDED adaptation.

\section{Conclusion}
\label{sec:conclusion}

We studied sharded multi-turn task solving as a self-generated-history problem:
when requirements arrive across turns, intermediate assistant replies enter
later context and can hinder final completion. \method{} addresses this with
adaptive on-policy self-distillation on the student's own sharded rollouts,
combining history-cleaned middle-turn supervision, answer-turn recovery, and
full-prompt preservation. Across math, tool/action generation, SQL, and Python
code generation, \method{} improves SHARDED completion over SFT, sparse-reward
RL, and single-turn-anchor RL while keeping FULL performance comparable. These
results suggest that robust multi-turn solving benefits from supervising
student-generated histories under reference contexts that preserve
user-provided requirements while removing earlier assistant replies.

\section*{Limitations}
\method{} may improve a language model's ability to preserve and recover task requirements
over longer conversations, but the same capability could also make harmful or
inappropriate objectives more persistent. The reference views used during
training can also encode biased, incorrect, or unsafe requirements, and those
errors may be distilled into the student model. Our evaluation focuses on
objectively scored tasks, so it does not establish safety, privacy protection,
or refusal behavior in open-domain dialogue. Deployment should therefore pair
the method with safety filtering, permission controls, and careful handling of
sensitive information.

\section*{Acknowledgments}
This work was supported by the NSFC (62436007), the Zhejiang NSF
(LZ25F020004), the Key R\&D Program in Zhejiang Province (2025C01128).

\bibliography{custom}

@inproceedings{NEURIPS2022_b1efde53,
 author = {Ouyang, Long and Wu, Jeffrey and Jiang, Xu and Almeida, Diogo and Wainwright, Carroll and Mishkin, Pamela and Zhang, Chong and Agarwal, Sandhini and Slama, Katarina and Ray, Alex and Schulman, John and Hilton, Jacob and Kelton, Fraser and Miller, Luke and Simens, Maddie and Askell, Amanda and Welinder, Peter and Christiano, Paul F and Leike, Jan and Lowe, Ryan},
 booktitle = {Advances in Neural Information Processing Systems},
 editor = {S. Koyejo and S. Mohamed and A. Agarwal and D. Belgrave and K. Cho and A. Oh},
 pages = {27730--27744},
 publisher = {Curran Associates, Inc.},
 title = {Training language models to follow instructions with human feedback},
 url = {https://proceedings.neurips.cc/paper_files/paper/2022/file/b1efde53be364a73914f58805a001731-Paper-Conference.pdf},
 volume = {35},
 year = {2022}
}

@article{grattafiori2024llama,
  title={The {Llama 3} herd of models},
  author={Grattafiori, Aaron and Dubey, Abhimanyu and Jauhri, Abhinav and Pandey, Abhinav and Kadian, Abhishek and Al-Dahle, Ahmad and Letman, Aiesha and Mathur, Akhil and Schelten, Alan and Vaughan, Alex and others},
  journal={arXiv preprint arXiv:2407.21783},
  year={2024}
}

@article{liu2024lost,
  title={Lost in the middle: How language models use long contexts},
  author={Liu, Nelson F and Lin, Kevin and Hewitt, John and Paranjape, Ashwin and Bevilacqua, Michele and Petroni, Fabio and Liang, Percy},
  journal={Transactions of the association for computational linguistics},
  volume={12},
  pages={157--173},
  year={2024},
  address={Cambridge, MA},
  publisher={MIT Press},
  doi={10.1162/tacl_a_00638}
}

@inproceedings{ross2011reduction,
  title={A reduction of imitation learning and structured prediction to no-regret online learning},
  author={Ross, St{\'e}phane and Gordon, Geoffrey and Bagnell, Drew},
  booktitle={Proceedings of the Fourteenth International Conference on Artificial Intelligence and Statistics},
  pages={627--635},
  year={2011},
  volume={15},
  series={Proceedings of Machine Learning Research},
  address={Fort Lauderdale, FL, USA},
  publisher={PMLR}
}

@inproceedings{bengio2015scheduled,
  title={Scheduled sampling for sequence prediction with recurrent neural networks},
  author={Bengio, Samy and Vinyals, Oriol and Jaitly, Navdeep and Shazeer, Noam},
  booktitle={Advances in Neural Information Processing Systems},
  volume={28},
  publisher={Curran Associates, Inc.},
  year={2015}
}

@inproceedings{li2025verifiable,
  title={Mitigating Lost in Multi-turn Conversation via Curriculum {RL} with Verifiable Accuracy and Abstention Rewards},
  author={Li, Ming and Chen, Pei and Zhang, Zhenhao and Yang, Tao and Zhang, Xinyang and Li, Han and Cao, Tianyu and Zeng, Ming and Wu, Zhuofeng and Jiang, Meng and Li, Huasheng and Li, Lihong and Yin, Bing},
  booktitle={Proceedings of the 64th Annual Meeting of the Association for Computational Linguistics (Volume 1: Long Papers)},
  pages={33364--33381},
  year={2026},
  address={San Diego, California, United States},
  publisher={Association for Computational Linguistics},
  doi={10.18653/v1/2026.acl-long.1540}
}

@inproceedings{zhao2026self,
  title={Self-Distilled Reasoner: On-Policy Self-Distillation for Large Language Models},
  author={Zhao, Siyan and Xie, Zhihui and Liu, Mengchen and Huang, Jing and Pang, Guan and Chen, Feiyu and Grover, Aditya},
  booktitle={International Conference on Machine Learning},
  year={2026},
  url={https://openreview.net/forum?id=Jpxfof0EaS}
}

@inproceedings{chen2026breaking,
  title={Breaking Contextual Inertia: Reinforcement Learning with Single-Turn Anchors for Stable Multi-Turn Interaction},
  author={Chen, Xingwu and Zhang, Zhanqiu and Guo, Steven Y. and Zou, Difan},
  booktitle={Findings of the Association for Computational Linguistics: ACL 2026},
  pages={6302--6318},
  year={2026},
  address={San Diego, California, United States},
  publisher={Association for Computational Linguistics},
  doi={10.18653/v1/2026.findings-acl.313}
}

@inproceedings{qiu2026pause,
  title={Pause or Fabricate? Training Language Models for Grounded Reasoning},
  author={Qiu, Yiwen and Wu, Linjuan and Liu, Yizhou and Yan, Yuchen and Ma, Jin and Tan, Xu and Hu, Yao and Zhang, Daoxin and Zhang, Wenqi and Lu, Weiming and Xiao, Jun and Shen, Yongliang},
  booktitle={Findings of the Association for Computational Linguistics: ACL 2026},
  pages={23056--23077},
  year={2026},
  address={San Diego, California, United States},
  publisher={Association for Computational Linguistics},
  doi={10.18653/v1/2026.findings-acl.1155}
}

@article{zheng2026useless,
  title={Useless but Safe? Benchmarking Utility Recovery with User Intent Clarification in Multi-Turn Conversations},
  author={Zheng, Mingqian and Morgan, Malia and Jiang, Liwei and Rose, Carolyn and Sap, Maarten},
  journal={arXiv preprint arXiv:2604.27093},
  year={2026}
}

@inproceedings{zheng2026pilot,
  title={{PILOT}: Planning via Internalized Latent Optimization Trajectories for Large Language Models},
  author={Zheng, Haoyu and Zhu, Yun and Yuan, Yuqian and Yuan, Bo and Zhang, Wenqiao and Tang, Siliang and Xiao, Jun},
  booktitle={Proceedings of the 64th Annual Meeting of the Association for Computational Linguistics (Volume 1: Long Papers)},
  pages={5204--5222},
  year={2026},
  address={San Diego, California, United States},
  publisher={Association for Computational Linguistics},
  doi={10.18653/v1/2026.acl-long.236}
}

@article{kruthof2026models,
  title={Models Recall What They Violate: Constraint Adherence in Multi-Turn {LLM} Ideation},
  author={Kruthof, Garvin},
  journal={arXiv preprint arXiv:2604.28031},
  year={2026}
}

@article{dongre2026attention,
  title={When Attention Closes: How {LLMs} Lose the Thread in Multi-Turn Interaction},
  author={Dongre, Vardhan and Hsieh, Joseph and Lai, Viet Dac and Yoon, Seunghyun and Bui, Trung and Hakkani-T{\"u}r, Dilek},
  journal={arXiv preprint arXiv:2605.12922},
  year={2026}
}

@article{he2026grounded,
  title={Grounded Continuation: A Linear-Time Runtime Verifier for {LLM} Conversations},
  author={He, Qisong and Dong, Yi and Huang, Xiaowei},
  journal={arXiv preprint arXiv:2605.14175},
  year={2026}
}

@inproceedings{laban2025llms,
  title={{LLMs} get lost in multi-turn conversation},
  author={Laban, Philippe and Hayashi, Hiroaki and Zhou, Yingbo and Neville, Jennifer},
  booktitle={International Conference on Learning Representations},
  year={2026},
  url={https://openreview.net/forum?id=VKGTGGcwl6}
}

@article{li2022alphacode,
  title={Competition-Level Code Generation with {A}lpha{C}ode},
  author={Li, Yujia and Choi, David and Chung, Junyoung and Kushman, Nate and Schrittwieser, Julian and Leblond, R{\'e}mi and Eccles, Tom and Keeling, James and Gimeno, Felix and Dal Lago, Agustin and Hubert, Thomas and Choy, Peter and de Masson d'Autume, Cyprien and Babuschkin, Igor and Chen, Xinyun and Huang, Po-Sen and Welbl, Johannes and Gowal, Sven and Cherepanov, Alexey and Molloy, James and Mankowitz, Daniel J. and Sutherland Robson, Esme and Kohli, Pushmeet and de Freitas, Nando and Kavukcuoglu, Koray and Vinyals, Oriol},
  journal={Science},
  volume={378},
  number={6624},
  pages={1092--1097},
  year={2022},
  doi={10.1126/science.abq1158}
}

@inproceedings{hendrycks2021apps,
  title={Measuring Coding Challenge Competence With {APPS}},
  author={Hendrycks, Dan and Basart, Steven and Kadavath, Saurav and Mazeika, Mantas and Arora, Akul and Guo, Ethan and Burns, Collin and Puranik, Samir and He, Horace and Song, Dawn and Steinhardt, Jacob},
  booktitle={Proceedings of the Neural Information Processing Systems Track on Datasets and Benchmarks},
  volume={1},
  year={2021},
  url={https://datasets-benchmarks-proceedings.neurips.cc/paper_files/paper/2021/file/c24cd76e1ce41366a4bbe8a49b02a028-Paper-round2.pdf}
}

@article{kang2026orderprobe,
  title={How Order-Sensitive Are {LLMs}? {O}rder{P}robe for Deterministic Structural Reconstruction},
  author={Kang, Zhaolu and He, Yingjie and Jiang, Kehan and Zheng, Leqi and Qian, Jiachen and Zhang, Qianyuan and Meng, Chunlei and Feng, Yujie and Wang, Yuan and Dou, Stephen and Wu, Aming and Zhao, Pengxiang and Liu, Jiaxin and Wang, Guansu and Zhang, Zeyu and Wang, Lei and Zhan, Qishi and He, Xiaomin and Zhang, Meisheng and Ni, Jianyuan and Xuan, Richeng},
  journal={arXiv preprint arXiv:2601.08626},
  year={2026},
  note={Findings of EMNLP 2026}
}

@article{shi2026spader,
  title={{SPADER}: Step-wise Peer Advantage with Diversity-Aware Exploration Rewards for Multi-Answer Question Answering},
  author={Shi, Qiming and Kang, Zhaolu and Zhou, Yunfan and Weng, Di and Wu, Yingcai},
  journal={arXiv preprint arXiv:2606.00593},
  year={2026}
}

@inproceedings{wang2025mucar,
  title={{MUCAR}: Benchmarking Multilingual Cross-Modal Ambiguity Resolution for Multimodal Large Language Models},
  author={Wang, Xiaolong and Kang, Zhaolu and Zhai, Wangyuxuan and Lou, Xinyue and Lai, Yunghwei and Wang, Ziyue and Wang, Yawen and Huang, Kaiyu and Wang, Yile and Li, Peng and Liu, Yang},
  booktitle={Proceedings of the 2025 Conference on Empirical Methods in Natural Language Processing},
  pages={15026--15048},
  year={2025},
  address={Suzhou, China},
  publisher={Association for Computational Linguistics},
  doi={10.18653/v1/2025.emnlp-main.760}
}

@inproceedings{weihua2026adamcot,
  title={{AdaMCoT}: Rethinking Cross-Lingual Factual Reasoning Through Adaptive Multilingual Chain-of-Thought},
  author={Zheng, Weihua and Huang, Xin and Liu, Zhengyuan and Vangani, Tarun Kumar and Zou, Bowei and Tao, Xiyan and Wu, Yuhao and Aw, AiTi and Chen, Nancy F. and Lee, Roy Ka-Wei},
  booktitle={Proceedings of the AAAI Conference on Artificial Intelligence},
  volume={40},
  pages={33863--33871},
  year={2026},
  doi={10.1609/aaai.v40i40.40678}
}

@inproceedings{weihua2025ccl,
  title={{CCL-XCoT}: An Efficient Cross-Lingual Knowledge Transfer Method for Mitigating Hallucination Generation},
  author={Zheng, Weihua and Lee, Roy Ka-Wei and Liu, Zhengyuan and Wu, Kui and Aw, AiTi and Zou, Bowei},
  booktitle={Findings of the Association for Computational Linguistics: EMNLP 2025},
  pages={1768--1788},
  year={2025},
  address={Suzhou, China},
  publisher={Association for Computational Linguistics},
  doi={10.18653/v1/2025.findings-emnlp.93}
}

@inproceedings{zheng2026mmac,
  title={{MMAC}: A Multilingual, Multimodal Alignment Framework for Cultural Grounding Evaluation},
  author={Zheng, Weihua and Liu, Zhengyuan and Chakraborty, Tanmoy and Xu, Weiwen and Gao, Xiaoxue and Tan, Bryan Chen Zhengyu and Zou, Bowei and Liu, Chang and Hu, Yujia and Xie, Xing and Yi, Xiaoyuan and Yao, Jing and Wang, Chaojun and Li, Long and Liu, Rui and Liu, Huiyao and Inoue, Koji and Sumida, Ryuichi and Kawahara, Tatsuya and Xu, Fan and Ye, Lingyu and Tian, Wei and Kim, Dongjun and Jung, Jimin and Seo, Jaehyung and Wangsajaya, Nadya Yuki and Duc, Pham Minh and Saxena, Ojasva and Nandi, Palash and Tao, Xiyan and Karlina, Wiwik and Luong, Tuan and Vasan, Keertana Arun and Lee, Roy Ka-Wei and Chen, Nancy F.},
  booktitle={Proceedings of the 64th Annual Meeting of the Association for Computational Linguistics (Volume 1: Long Papers)},
  pages={21667--21709},
  year={2026},
  address={San Diego, California, United States},
  publisher={Association for Computational Linguistics},
  doi={10.18653/v1/2026.acl-long.989}
}

@article{canaverde2026sequor,
  title={{SEQUOR}: A Multi-Turn Benchmark for Realistic Constraint Following},
  author={Canaverde, Beatriz and Alves, Duarte M and Pombal, Jos{\'e} and Attanasio, Giuseppe and Martins, Andr{\'e} FT},
  journal={arXiv preprint arXiv:2605.06353},
  year={2026}
}

@inproceedings{atri2026evaluating,
  title={Evaluating Temporal Consistency in Multi-Turn Language Models},
  author={Atri, Yash Kumar and Johnson, Steven L. and Hartvigsen, Thomas},
  booktitle={Proceedings of the 64th Annual Meeting of the Association for Computational Linguistics (Volume 1: Long Papers)},
  pages={45964--45982},
  year={2026},
  address={San Diego, California, United States},
  publisher={Association for Computational Linguistics},
  doi={10.18653/v1/2026.acl-long.2133}
}

@inproceedings{wan2026mitigating,
  title={Mitigating Conversational Inertia in Multi-Turn Agents},
  author={Wan, Yang and Cao, Zheng and Zhang, Zhenhao and Zeng, Zhengwen and Shen, Shuheng and Meng, Changhua and Zhu, Linchao},
  booktitle={International Conference on Machine Learning},
  year={2026},
  url={https://openreview.net/forum?id=TCWXiFT7gj}
}

@inproceedings{agarwal2024policy,
  title={On-policy distillation of language models: Learning from self-generated mistakes},
  author={Agarwal, Rishabh and Vieillard, Nino and Zhou, Yongchao and Stanczyk, Piotr and Ramos Garea, Sabela and Geist, Matthieu and Bachem, Olivier},
  booktitle={International Conference on Learning Representations},
  year={2024},
  url={https://openreview.net/forum?id=3zKtaqxLhW}
}

@article{ye2026policy,
  title={On-policy context distillation for language models},
  author={Ye, Tianzhu and Dong, Li and Wu, Xun and Huang, Shaohan and Wei, Furu},
  journal={arXiv preprint arXiv:2602.12275},
  year={2026}
}

@article{zhang2026opsdl,
  title={{OPSDL}: On-policy self-distillation for long-context language models},
  author={Zhang, Xinsen and Ding, Zhenkai and Pan, Tianjun and Yang, Run and Kang, Chun and Xiong, Xue and Gu, Jingnan},
  journal={arXiv preprint arXiv:2604.17535},
  year={2026}
}

@article{zhang2026shortopd,
  title={{ShortOPD}: Recovering Pruned {LLMs} with Short-to-Long On-Policy Distillation},
  author={Zhang, Qingyu and Yuan, Qianhao and Lin, Hongyu and Lu, Yaojie and Han, Xianpei and Sun, Le and Xu, Ming and Li, Jiarui},
  journal={arXiv preprint arXiv:2607.13124},
  year={2026}
}

@article{yuan2026visionopd,
  title={{Vision-OPD}: Learning to See Fine Details for Multimodal {LLMs} via On-Policy Self-Distillation},
  author={Yuan, Qianhao and Lou, Jie and Yu, Xing and Lin, Hongyu and Sun, Le and Han, Xianpei and Lu, Yaojie},
  journal={arXiv preprint arXiv:2605.18740},
  year={2026}
}

@article{zhu2026many,
  title={The Many Faces of On-Policy Distillation: Pitfalls, Mechanisms, and Fixes},
  author={Zhu, Siqi and Ye, Xuyan and Lu, Hongyu and Shi, Weiye and Liu, Ge},
  journal={arXiv preprint arXiv:2605.11182},
  year={2026}
}

@article{yu2026multi,
  title={Multi-Rollout On-Policy Distillation via Peer Successes and Failures},
  author={Yu, Weichen and Li, Xiaomin and Zhao, Yizhou and Liu, Xiaoze and Zhang, Ruowang and Wang, Haixin and Luo, Yinyi and Wu, Chen Henry and Mittal, Gaurav and Fredrikson, Matt and others},
  journal={arXiv preprint arXiv:2605.12652},
  year={2026}
}

@article{wang2026tcod,
  title={{TCOD}: Exploring Temporal Curriculum in On-Policy Distillation for Multi-turn Autonomous Agents},
  author={Wang, Jiaqi and Zhang, Wenhao and Shi, Weijie and Li, Yaliang and Cheng, James},
  journal={arXiv preprint arXiv:2604.24005},
  year={2026}
}

@article{zheng2026scope,
  title={{SCOPE}: Signal-calibrated on-policy distillation enhancement with dual-path adaptive weighting},
  author={Zheng, Binbin and Ma, Xing and Liang, Yiheng and Ruan, Jingqing and Fu, Xiaoliang and Lin, Kepeng and Zhu, Benchang and Zeng, Ke and Cai, Xunliang},
  journal={arXiv preprint arXiv:2604.10688},
  year={2026}
}

@article{liu2026intent,
  title={Intent mismatch causes {LLMs} to get lost in multi-turn conversation},
  author={Liu, Geng and Zhu, Fei and Feng, Rong and Ma, Changyi and Wang, Shiqi and Meng, Gaofeng},
  journal={arXiv preprint arXiv:2602.07338},
  year={2026}
}

@inproceedings{wang2026icpo,
  title={{ICPO}: Illocution-Calibrated Policy Optimization for Multi-Turn Conversation},
  author={Wang, Zhebo and Mu, Xiaohu and Zhou, Zijie and Li, Mohan and Xing, Wenpeng and Kong, Dezhang and Han, Meng},
  booktitle={ICASSP 2026--2026 IEEE International Conference on Acoustics, Speech and Signal Processing (ICASSP)},
  pages={16917--16921},
  year={2026},
  organization={IEEE},
  doi={10.1109/ICASSP55912.2026.11462691}
}

@article{shao2024deepseekmath,
  title={{DeepSeekMath}: Pushing the limits of mathematical reasoning in open language models},
  author={Shao, Zhihong and Wang, Peiyi and Zhu, Qihao and Xu, Runxin and Song, Junxiao and Bi, Xiao and Zhang, Haowei and Zhang, Mingchuan and Li, YK and Wu, Yang and others},
  journal={arXiv preprint arXiv:2402.03300},
  year={2024}
}

@article{liu2024deepseek,
  title={{DeepSeek-V3} technical report},
  author={{DeepSeek-AI} and others},
  journal={arXiv preprint arXiv:2412.19437},
  year={2025}
}

@article{yang2025qwen3,
  title={{Qwen3} technical report},
  author={{Qwen Team}},
  journal={arXiv preprint arXiv:2505.09388},
  year={2025}
}

@article{cobbe2021training,
  title={Training verifiers to solve math word problems},
  author={Cobbe, Karl and Kosaraju, Vineet and Bavarian, Mohammad and Chen, Mark and Jun, Heewoo and Kaiser, Lukasz and Plappert, Matthias and Tworek, Jerry and Hilton, Jacob and Nakano, Reiichiro and others},
  journal={arXiv preprint arXiv:2110.14168},
  year={2021}
}

@inproceedings{yu2018spider,
  title={Spider: A large-scale human-labeled dataset for complex and cross-domain semantic parsing and text-to-{SQL} task},
  author={Yu, Tao and Zhang, Rui and Yang, Kai and Yasunaga, Michihiro and Wang, Dongxu and Li, Zifan and Ma, James and Li, Irene and Yao, Qingning and Roman, Shanelle and Zhang, Zilin and Radev, Dragomir},
  booktitle={Proceedings of the 2018 Conference on Empirical Methods in Natural Language Processing},
  pages={3911--3921},
  year={2018},
  address={Brussels, Belgium},
  publisher={Association for Computational Linguistics},
  doi={10.18653/v1/D18-1425}
}

@article{chen2021evaluating,
  title={Evaluating large language models trained on code},
  author={Chen, Mark and Tworek, Jerry and Jun, Heewoo and Yuan, Qiming and Pinto, Henrique Ponde De Oliveira and Kaplan, Jared and Edwards, Harri and Burda, Yuri and Joseph, Nicholas and Brockman, Greg and others},
  journal={arXiv preprint arXiv:2107.03374},
  year={2021}
}

@inproceedings{patil2025berkeley,
  title={The {Berkeley} function calling leaderboard ({BFCL}): From tool use to agentic evaluation of large language models},
  author={Patil, Shishir G and Mao, Huanzhi and Yan, Fanjia and Ji, Charlie Cheng-Jie and Suresh, Vishnu and Stoica, Ion and Gonzalez, Joseph E},
  booktitle={Proceedings of the 42nd International Conference on Machine Learning},
  pages={48371--48392},
  year={2025},
  volume={267},
  series={Proceedings of Machine Learning Research},
  publisher={PMLR}
}

@article{qwen2025qwen25technicalreport,
  title={{Qwen2.5} Technical Report},
  author={{Qwen Team}},
  journal={arXiv preprint arXiv:2412.15115},
  year={2025}
}

@inproceedings{hu2022lora,
  title={{LoRA}: Low-rank adaptation of large language models},
  author={Hu, Edward J. and Shen, Yelong and Wallis, Phillip and Allen-Zhu, Zeyuan and Li, Yuanzhi and Wang, Shean and Wang, Lu and Chen, Weizhu},
  booktitle={International Conference on Learning Representations},
  year={2022},
  url={https://openreview.net/forum?id=nZeVKeeFYf9}
}

\clearpage
\appendix
\makeatletter
\renewcommand{\@seccntformat}[1]{
  \@ifundefined{appendix@#1@format}
    {\csname the#1\endcsname\quad}
    {\csname appendix@#1@format\endcsname}}
\newcommand{\appendix@section@format}{Appendix~\thesection.\quad}
\makeatother

\section{Experimental Protocol and Reproducibility}
\label{sec:appendix-exp-details}

\definecolor{maigogray}{HTML}{F7F7F7}
\definecolor{maigomid}{HTML}{7A5B00}
\definecolor{maigogood}{HTML}{006B3F}
\definecolor{maigobad}{HTML}{9A1C1C}

\lstdefinestyle{maigoprompt}{
    basicstyle=\ttfamily\scriptsize,
    breaklines=true,
    breakatwhitespace=false,
    columns=fullflexible,
    keepspaces=true,
    frame=single,
    rulecolor=\color{black!30},
    xleftmargin=0.5em,
    xrightmargin=0.5em,
    aboveskip=0.4em,
    belowskip=0.4em
}
\lstdefinestyle{maigoprompttiny}{
    basicstyle=\ttfamily\scriptsize,
    breaklines=true,
    breakatwhitespace=false,
    columns=fullflexible,
    keepspaces=true,
    frame=single,
    rulecolor=\color{black!30},
    xleftmargin=0.25em,
    xrightmargin=0.25em,
    aboveskip=0.25em,
    belowskip=0.25em
}
\lstdefinestyle{maigocase}{
    basicstyle=\ttfamily\scriptsize,
    breaklines=true,
    breakatwhitespace=false,
    columns=fullflexible,
    keepspaces=true,
    frame=none,
    backgroundcolor=\color{maigogray},
    xleftmargin=0.25em,
    xrightmargin=0.25em,
    aboveskip=0.25em,
    belowskip=0.25em
}

\newcommand{\badtag}[1]{\textcolor{maigobad}{\textbf{#1}}}
\newcommand{\goodtag}[1]{\textcolor{maigogood}{\textbf{#1}}}
\newcommand{\midtag}[1]{\textcolor{maigomid}{\textbf{#1}}}
\newcommand{\promptheading}[1]{
  \par\bigskip
  \noindent\begingroup
  \setlength{\fboxsep}{5pt}
  \colorbox{black!55}{\parbox{\dimexpr\linewidth-2\fboxsep-6\fboxrule\relax}{\color{white}\bfseries\small #1}}
  \endgroup\par\nobreak\vspace{0.35em}
}
\newcommand{\promptfield}[1]{\par\vspace{0.28em}\noindent\textbf{#1}\par}
\newcommand{\codeblock}[1]{
  \vspace{0.15em}\noindent\fbox{\begin{minipage}{0.965\linewidth}\scriptsize\ttfamily #1\end{minipage}}\par\vspace{0.15em}
}
\newcommand{\promptend}{\par\smallskip\noindent\hrule\par\medskip}
\newcommand{\caseline}[3]{
  \par\noindent\hangindent=1.9em\hangafter=1
  \textbf{#1 #2:} #3\par}
\newcommand{\caseuser}[2]{\caseline{User}{#1}{#2}}
\newcommand{\caseassistant}[2]{\caseline{Assistant}{#1}{#2}}
\newcommand{\caseassistantblock}[1]{\par\noindent\textbf{Assistant #1:}\par\nobreak}
\newcommand{\caseresult}[1]{\par\noindent\textbf{Result:} #1\par}

This appendix is organized around the information needed to audit and
reproduce the experiments. Appendix~\ref{sec:appendix-tasks} describes the
paired task construction, Appendix~\ref{sec:appendix-optimization} gives the
training settings, Appendix~\ref{sec:appendix-reference-views} specifies the
training-only reference contexts, Appendix~\ref{sec:appendix-prompts} records
the prompt templates, and Appendix~\ref{sec:appendix-cases} provides
qualitative multi-turn cases.

\subsection{Tasks and Metrics}
\label{sec:appendix-tasks}

Each task is evaluated in paired FULL and SHARDED views: FULL presents the
complete task in one prompt, while SHARDED reveals the same requirements across
user turns and scores the final answer. The evaluation uses GSM8K
(\(N{=}1319\), numeric answer match), Actions (\(N{=}382\), tool-call AST
match), Database (\(N{=}715\), SQL execution match), and HumanEval
(\(N{=}164\), unit tests).

Table~\ref{tab:appendix-construction} summarizes how the paired views are
constructed. We follow the LiC principle that FULL and SHARDED expose the same
task requirements, with SHARDED delaying their presentation across user turns.
The conversion follows the fixed task-family protocols used for our LiC-style
evaluation and is shared by all methods; the table reports the information
needed to interpret the comparison at a protocol level.

\begin{table*}[t]
\centering
\footnotesize
\setlength{\tabcolsep}{3pt}
\renewcommand{\arraystretch}{1.08}
\begin{tabular}{@{}>{\raggedright\arraybackslash}p{0.11\textwidth}
>{\raggedright\arraybackslash}p{0.18\textwidth}
>{\raggedright\arraybackslash}p{0.24\textwidth}
>{\raggedright\arraybackslash}p{0.25\textwidth}
>{\raggedright\arraybackslash}p{0.13\textwidth}@{}}
\toprule
Task & Fine-tuning set & Evaluation split & SHARDED construction & Scoring note \\
\midrule
GSM8K & 6,400 GSM8K training tasks converted to sharded math dialogues
& Full held-out GSM8K/LiC-style math set; 1,319 matched examples
& LiC-style math sharding reveals the requested quantity and supporting facts
across turns.
& Numeric answer match. \\
Actions & 6,400 BFCL-derived training tasks converted to sharded tool-use
dialogues
& BFCL-derived held-out parallel and parallel-multiple items; 382 matched
examples
& Tool-use requirements are exposed conversationally under the same
tool-call format used by FULL.
& Strict AST match. \\
Database & 6,397 Spider training tasks converted to sharded text-to-SQL dialogues
& Spider dev examples under the same fixed conversion; 715 matched examples
& The database schema is fixed in the system context, while the
natural-language query and constraints are sharded.
& SQL execution match; fixed-schema sharded text-to-SQL. \\
HumanEval & 6,400 Python code-generation training tasks converted to sharded
specification dialogues. The pool contains 3,621 filtered DeepMind Code
Contests examples as the primary source and 2,779 deduplicated APPS training
examples as fill~\citep{li2022alphacode,hendrycks2021apps}
& Full HumanEval split through the Python function-generation harness; 164
matched examples
& Function requirements are revealed through short user turns under a fixed
sharded code-specification protocol.
& Unit-test pass@1. \\
\bottomrule
\end{tabular}
\caption{Task construction for paired FULL/SHARDED evaluation.}
\label{tab:appendix-construction}
\end{table*}

For SHARDED interaction, we retain the simulator roles from the LiC protocol: an evaluated
assistant, a model-assisted user model that decides which unrevealed shard to
reveal next, and a system component that classifies answer attempts and extracts
the candidate answer for the task verifier. Our experiments use a fixed
\deepseekvthree{} model for the user model, answer-attempt classifier, and
answer extractor. This model controls only the interaction environment around the
evaluated assistant; the sharding protocol, turn budget, answer extraction
interface, and deterministic task verifiers remain aligned with the LiC
evaluation protocol. Every model and training method is evaluated with the same
simulator model, same shard files, same prompt templates, same turn budget, and
same task verifier.

Outputs that fail to parse, are empty, violate the response protocol, exceed
the context limit, or reach the interaction limit are counted as incorrect.
This convention keeps FULL and SHARDED comparable: a model must both produce an
answer in the expected format and solve the underlying task.

\paragraph{Evaluation decoding and failure handling.}
The evaluated assistant and simulated user both use temperature 1.0. The
answer-attempt classifier and answer extractor use temperature 0.0. Assistant
outputs are capped at 2,048 tokens for non-code tasks and 4,096 tokens for
Code, under a 32,768-token context window. Structured simulator outputs receive
up to four schema attempts, and answer extraction receives up to three
attempts. Malformed outputs, timeouts, protocol violations, exhausted retries,
and evaluation errors are all counted as incorrect.

\subsection{Optimization and Baselines}
\label{sec:appendix-optimization}

All trained methods are optimized as LoRA adapters~\citep{hu2022lora} and
evaluated after merging the adapter weights into the base model.
Tables~\ref{tab:appendix-train-settings}
and~\ref{tab:appendix-maigo-settings} summarize the optimization budget and
\method{}-specific settings used in the main experiments.

\begin{table*}[t]
\centering
\small
\setlength{\tabcolsep}{4pt}
\renewcommand{\arraystretch}{1.08}
\begin{tabular}{@{}>{\raggedright\arraybackslash}p{0.20\textwidth}
>{\raggedright\arraybackslash}p{0.24\textwidth}
>{\raggedright\arraybackslash}p{0.24\textwidth}
>{\raggedright\arraybackslash}p{0.24\textwidth}@{}}
\toprule
Setting & \method{} & SFT & GRPO / RLSTA \\
\midrule
Adapter & LoRA rank 64, alpha 128; attention and MLP projections
& Same as \method{} & Same as \method{} \\
Updates & 100 optimizer steps & 100 optimizer steps & 100 optimizer steps \\
Hardware and dtype & Same hardware configuration, bfloat16 & Same hardware configuration, bfloat16 & Same hardware configuration, bfloat16 \\
Learning rate & \(5\times10^{-6}\) & \(5\times10^{-6}\) & \(3\times10^{-7}\) \\
Batching & Per-device batch 4; grad. accumulation 1; global batch 64
& Per-device batch 4; grad. accumulation 1
& Per-device prompt batch 8; eight generations per prompt \\
Training context & 16,384 tokens & 16,384 tokens & 16,384 tokens \\
Evaluation context & \multicolumn{3}{l}{32,768 tokens for all FULL and SHARDED evaluations} \\
\bottomrule
\end{tabular}
\caption{Training and evaluation settings for fine-tuning comparisons.}
\label{tab:appendix-train-settings}
\end{table*}

\begin{table}[t]
\centering
\footnotesize
\setlength{\tabcolsep}{4pt}
\renewcommand{\arraystretch}{1.08}
\begin{tabular}{@{}>{\raggedright\arraybackslash}p{0.42\columnwidth}
>{\raggedright\arraybackslash}p{0.50\columnwidth}@{}}
\toprule
\method{} setting & Value \\
\midrule
Rollout sampling & temperature 1.0, top-\(p=0.95\), top-\(k=20\) \\
Final/full generation budget & 4,096 tokens during training \\
Middle-turn generation budget & 256 tokens during training \\
Reference update & EMA reference policy, decay 0.99 \\
Final/FULL loss parameter & \(\beta_{\mathrm{final}}=\beta_{\mathrm{full}}=1.0\) \\
Middle loss parameter & \(\beta_{\mathrm{mid}}=0.5\) \\
Middle-loss scale & \(\alpha_t=1.0\) for all eligible turns \\
Full-branch rate & \(\rho=0.2\) \\
Token-level clipping & 0.5 \\
\bottomrule
\end{tabular}
\caption{\method{} rollout and objective hyperparameters.}
\label{tab:appendix-maigo-settings}
\end{table}

\begin{table}[t]
\centering
\footnotesize
\setlength{\tabcolsep}{4pt}
\renewcommand{\arraystretch}{1.08}
\begin{tabular}{@{}lrrrr@{}}
\toprule
Task & Windows & \(\overline{w_t}\) & Range of \(w_t\) & \(\overline{\delta_t}\) \\
\midrule
Actions & 99 & 0.517 & 0.000--1.000 & 0.122 \\
Database & 99 & 0.489 & 0.129--0.923 & 0.309 \\
GSM8K & 99 & 0.427 & 0.033--0.828 & 0.233 \\
HumanEval & 99 & 0.486 & 0.155--0.817 & 0.257 \\
\bottomrule
\end{tabular}
\caption{Training-time diagnostics for the sampled adaptive middle-turn weight
\(w_t\) in 7B \method{} runs, measured from one training seed.}
\label{tab:appendix-mid-weight-stats}
\end{table}

Table~\ref{tab:appendix-mid-weight-stats} reports a descriptive training
diagnostic rather than an additional multi-seed comparison. The values are the
actual sampled middle-turn weights used by the optimizer. Across tasks, the
mean weight stays well below a uniform weight of one while remaining far from
zero, indicating that the adaptive rule changes the middle-turn signal without
collapsing supervision. The causal comparison for this component is the corresponding
uniform-weight ablation in the main ablation table.

The main comparison rows are matched fine-tuning comparisons under a common
adapter, training set construction, optimization budget, and evaluation setup.
The RL baselines use their own sparse task rewards; RLSTA additionally uses the
paired FULL view as its single-turn anchor reward. No method receives extra
evaluation-time state or a different SHARDED prompt.

\subsection{Training-Only Reference Views}
\label{sec:appendix-reference-views}

The reference contexts are used only for computing distillation losses during
training. They do not change the information available to the deployed student,
and they are built from the same paired task views used by the LiC-style
FULL/SHARDED protocol. The following rules define the three reference views used
by \method{}.

\paragraph{Middle turns.}
The middle-turn reference keeps the user turns through the current shard and
task-specific system information that is shared by the paired views. It removes
earlier assistant replies and withholds both unrevealed user turns and the full
task view.

\paragraph{Answer turn.}
The answer-turn reference keeps all user turns in order and preserves the same
task-specific system information as the evaluation prompt. It adds a canonical
serialization built from the completed dialogue and its paired FULL view, then
removes earlier assistant replies such as premature calls, draft SQL, or
partial code attempts. The paired FULL view is used only to render the
completed requirements in the fixed task format, not to add a gold answer or a
verifier signal.

\paragraph{Full branch.}
The full branch uses the complete single-turn task prompt and no sharded
history. It is an EMA anchor for FULL preservation rather than a source of new
sharded supervision.

The task-specific keep/remove rules are:
\begin{description}
\item[GSM8K.]
Middle references keep the math system instruction and user shards through
turn \(t\), while removing earlier assistant replies, later facts, and the full
problem text. Answer references keep all user shards and append a canonical
serialization of the completed problem requirements.
\item[Actions.]
Middle references keep the tool-use instruction and the user-visible tool
requirements revealed so far. Answer references keep all user turns and append
the completed tool-use request in canonical form, while removing earlier
assistant calls.
\item[Database.]
The schema remains in the system context for both student and reference.
Middle references keep query shards through turn \(t\); answer references keep
all query shards and append the completed natural-language query in canonical
form, while removing earlier draft SQL.
\item[HumanEval.]
Middle references keep the coding instruction and function-specification shards
through turn \(t\). Answer references keep all specification shards and append
the completed programming task requirements in canonical form, while removing
earlier partial code.
\end{description}
Thus the reference view changes the serialization of completed user-provided
requirements for training; it does not add a gold-answer label and does not
change what the student receives at inference time.

\section{Prompt and Template Cards}
\label{sec:appendix-prompts}

This section records the prompt surfaces used by the experiments. Dynamic
fields are written as \texttt{[[FIELD]]}. Per-item fields such as tool
schemas, database schemas, user shards, and function headers are filled from
the fixed train/evaluation files. We separate assistant-visible task prompts
from the prompts used by the interaction environment. The environment prompts
are sent as user-role requests to the fixed simulator/classifier model and are
not visible to the evaluated assistant.

\subsection{Interaction Environment Prompts}
\label{sec:appendix-simulator-prompts}

The SHARDED environment follows the LiC interaction structure. The first user
turn is the first stored shard and does not call the user-simulator model. From
the second user turn onward, the simulator receives the current conversation,
the revealed shard list, and the unrevealed shard list.

\promptheading{Prompt Card B1: user-simulator template}
\begin{lstlisting}[style=maigoprompttiny]
You are simulating a user of an interactive LLM system (like ChatGPT).
The user is inherently lazy, and answers in short form, providing only minimal information to the system. You should not be proactive.

Here's the conversation so far:
[[CONVERSATION_SO_FAR]]

Here are the shards that have already been revealed:
[[SHARDS_REVEALED]]

Here are all the shards that have not been revealed yet:
[[SHARDS_NOT_REVEALED]]

You must generate a response to the conversation so far. Here are the rules:
- [Providing a Shard] You can reveal the content of a shard to the system in your response if it will help the system move closer to answering the problem. You should select the shard to reveal that is most "basic" and is the current most relevant shard.
- [One Shard at a Time] You should only reveal at most one shard at a time.
- [Reveal Entire Shard] If you reveal a shard, you must make sure to include *all the information in the shard*. For example, if the shard is "your symptoms are that you have a headache in the mornings", your response can't just be ``yeah I have headaches'', you must say ``yup mostly headaches in the mornings``.
- [Irrelevant Clarifications] If the system asks you a question irrelevant to the shards, asks you a generic question (``Can you give me a hint?``), you should respond with an answer that does not provide a shard. (``I don't know``, ``Is that really important?``, etc.) You should not reveal any information beyond what is available in the shards.
- [No Repeated Shards] You should not reveal the same shard more than once. Carefully review the shards revealed already, and only reveal a shard if its `shard_id` is not on the list.
- [Rephrase Shards] If you reveal a shard, you should rephrase it in a conversational way. Do not copy the shard verbatim.
- [Do Not Ask Questions] Your response should always be declarative sentences, and not questions.
- [Brevity of Response] You should favor being succint. Your answer can also have typos, improper grammar, capitalization, etc. You are simulating a real person talking to an AI, who is in a hurry.
- [Format] Your response should be formatted as a JSON object with the following keys:
    - `response`: The response to the conversation so far.
    - `shard_id`: The shard you are revealing to the system. The shard_id can be an integer, or -1 if you did not reveal any shards.
For example:
{"response": "I don't know", "shard_id": -1}
or:
{"response": "yeah I want it to [...]", "shard_id": 1}
\end{lstlisting}

After each assistant response, a fixed classifier decides whether the turn is a
scorable answer attempt. Prompt Card B3 lists the task-specific answer surface
used by the classifier and extractor.

\promptheading{Prompt Card B2: answer-attempt classifier template}
\begin{lstlisting}[style=maigoprompttiny]
You are reviewing a multi-turn conversation between a user and an assistant, and are given the last turn of the conversation.

Here is the full specification of the problem the system is attempting to solve:
[[INITIAL_SHARD]]

Specification:
[[SHARDS]]

You must classify the response of the assistant according to the response type:
- `answer_attempt`: The response contains a complete answer attempt to the user's question (not templated or hypothetical), that can be extracted verbatim. See the task-specific answer description for more details.
- `clarification`: The response  is short (less than 100 words) and contains a single question addressed to the user that directly inquires about an aspect of the user's query. A clarification turn cannot be long (see `discussion`), cannot contain a vague question (see `discussion`) and cannot contain multiple questions (see `interrogation`).
- `interrogation`: The response contains multiple questions addressed to the user, sometimes organized in a list or bullet-points.
- `discussion`: The response discusses the question in detail, without providing a final answer, asking a specific clarification question, or a refusal to answer. The response may or may not contain a vague question (e.g., "What else can I help you with?").
- `hedge`: The response contains multiple answer candidates based on hypotheticals (ifs) or branching (case 1, case 2) with corresponding descriptions.
- `refuse`: The response contains an explicit or implicit refusal to answer the user's question without a follow-up question or a request.
- `missing`: The response is empty/blank.

You must output your answer in the following JSON format:
{"response_type": "refuse|missing|answer_attempt|hedge|clarification|interrogation|discussion"}

Rules:
- The assistant giving a hint at how an answer could look like is not a final answer.
- [Task Specific Answer] [[ANSWER_DESCRIPTION]]

Conversation's last turn:
[[CONVERSATION_SO_FAR]]
\end{lstlisting}

\promptheading{Prompt Card B3: answer surfaces}
\begingroup\small
\promptfield{Math}
The classifier looks for a single numeric answer; an LLM extractor returns the
exact numeric span for verifier matching.
\promptfield{Actions}
The full assistant response is scored as a BFCL-format function-call list.
\promptfield{Database}
The classifier looks for one complete executable SQL query; an LLM extractor
normalizes the submitted SQL span before execution scoring.
\promptfield{Code}
The task-specific Python extractor expects one valid submitted Python function.
\par\endgroup\promptend

\promptheading{Prompt Card B4: answer-span normalization}
\begingroup\small
For Math and Database, a fixed extraction prompt is used only to locate the
answer span before deterministic scoring. It does not judge correctness.
\promptfield{Math}
The extractor returns the exact numeric answer substring, which is then scored
by numeric matching.
\promptfield{Database}
The extractor returns a single SQL span without prose or code fences, which is
then scored by execution.
\par\endgroup\promptend

\subsection{Task System and FULL Prompt Templates}
\label{sec:appendix-task-prompts}

The evaluated assistant sees the following task system prompts. Actions fills
\texttt{[[FUNCTIONS]]} with the per-item BFCL function schema, and Database
fills \texttt{[[SCHEMA]]} with the Spider schema. For the main Code row,
HumanEval examples use the function-generation harness with the
Python-programmer system prompt and the HumanEval FULL user template shown
below.

\promptheading{Prompt Card B5: Math task prompts}
\begin{lstlisting}[style=maigoprompttiny]
System:
As an expert problem solver solve step by step the following mathematical questions.

FULL user:
Q: [[QUESTION]]
A:
\end{lstlisting}

\promptheading{Prompt Card B6: Actions task prompts}
\begin{lstlisting}[style=maigoprompttiny]
System:
You are an expert in composing functions. You are given a question and a set of possible functions. Based on the question, you will need to make one or more function/tool calls to achieve the purpose.
If none of the functions can be used, point it out. If the given question lacks the parameters required by the function, also point it out.
You should only return the function calls in your response.

If you decide to invoke any of the function(s), you MUST put it in the format of [func_name1(params_name1=params_value1, params_name2=params_value2...), func_name2(params)]
You SHOULD NOT include any other text in the response.

Here is a list of functions in JSON format that you can invoke.

[[FUNCTIONS]]

FULL user:
[[QUESTION]]
\end{lstlisting}

\promptheading{Prompt Card B7: Database task prompts}
\begin{lstlisting}[style=maigoprompttiny]
System:
You are helping a user write SQL queries to a database. If something is not clear, you can ask the user to clarify what they need. The schema for the database being accessed is the following:
        
[[SCHEMA]]

FULL user:
You must generate the valid SQL query that answers the given user query in natural language.

The schema of the database you are responding to is as follows:
[[DATABASE_SCHEMA]]

The user query is:
[[USER_QUERY]]

Rules:
- [Single Answer] Produce a single valid SQL query based on the database schema and the user query.
- [Only SQL] Do not include any other text or comments.
- [Validity] The SQL query must be valid and executable.
- [Single Line] Produce your entire response in a single line.
\end{lstlisting}

\promptheading{Prompt Card B8: Code task prompts}
\begin{lstlisting}[style=maigoprompttiny]
System:
You are an expert Python programmer. You will be given a question (problem specification) and will generate a correct Python program that matches the specification and passes all tests.

Format:
- [Standalone] Make sure that your answer consists of only one Python function at the top level. Do not wrap with a class or split into multiple functions.

FULL user:
You must generate a valid Python function that satisfies the following instruction:

Instruction:

Complete the following incomplete function signature:
```python
[[HUMANEVAL_PROMPT]]
```

Rules:
- [Format] You must only return the Python function, matching the signature exactly.
- [Validity] The Python function must be valid and executable. Do not include any other text or comments. (```python```)
\end{lstlisting}

\subsection{Evaluation and Training Message Forms}
\label{sec:appendix-eval-messages}

We write \(\langle x\rangle\) for the complete task view, \(\langle u_t\rangle\)
for the \(t\)-th user shard, \(\langle \hat a_t\rangle\) for a student-sampled
assistant reply, and \(\langle y^\star\rangle\) for the reference final answer
used by supervised baselines.

\promptheading{Prompt Card B9: FULL evaluation and FULL preservation}
\begin{lstlisting}[style=maigoprompt]
System: <task-specific system message from Appendix B.2>
User:   <complete task view x rendered with the task FULL template>
Assistant: <decoded answer, or sampled answer during FULL preservation>
\end{lstlisting}

\promptheading{Prompt Card B10: SHARDED evaluation}
\begin{lstlisting}[style=maigoprompt]
System: <task-specific system message from Appendix B.2>
User:   <u_1>
Assistant: <model reply a_1>
...
User:   <u_T>
Assistant: <final decoded answer>
\end{lstlisting}

\promptheading{Prompt Card B11: middle-turn distillation}
\begin{lstlisting}[style=maigoprompt]
Student context:
  System: <task-specific system message>
  User/Assistant history: <u_1, a_1, ..., u_{t-1}, a_{t-1}>
  User: <u_t>
  Assistant tokens used for loss: <sampled middle reply a_t>

Reference context:
  System: <same task-specific system message>
  User-only prefix: <u_1, ..., u_t>
  Assistant tokens used for loss: <the same sampled middle reply a_t>

Removed from the reference:
  previous assistant replies, unrevealed user turns, full task view.
\end{lstlisting}

\promptheading{Prompt Card B12: answer-turn recovery distillation}
\begin{lstlisting}[style=maigoprompt]
Student context:
  System: <task-specific system message>
  Full SHARDED history: <u_1, a_1, ..., u_{T-1}, a_{T-1}, u_T>
  Assistant tokens used for loss: <sampled final reply a_T>

Reference context:
  System: <same task-specific system message>
  User-only completed dialogue: <u_1, ..., u_T>
  Canonical task block: <completed requirements rendered from u_1, ..., u_T and x>
  Assistant tokens used for loss: <the same sampled final reply a_T>

Removed from the reference:
  previous assistant replies such as partial SQL, premature calls, or draft code.
\end{lstlisting}

\promptheading{Prompt Card B13: baseline sharded training messages}
\begin{lstlisting}[style=maigoprompt]
SFT:
  System: <task-specific system message>
  User: <u_1>
  ...
  User: <u_T>
  User: "Now answer the task using all information revealed so far.
         Return the final answer only in the task's expected format."
  Assistant target: <reference final answer y*>

GRPO / RLSTA:
  The policy uses the same SHARDED prompt as evaluation. Rewards score
  the decoded final answer; RLSTA additionally uses the paired FULL view
  as its single-turn anchor. Evaluation prompts are unchanged.
\end{lstlisting}

\subsection{Offline Shard Construction}
\label{sec:appendix-shard-prompts}

Shard construction is an offline preprocessing step. Runtime evaluation and
fine-tuning consume the resulting shard lists and do not call the shard
constructor. We use fixed task-family prompt programs following the LiC
paired-view protocol:
GSM8K uses arithmetic segmentation followed by conversational rewriting;
Actions uses BFCL request segmentation and conversational rewriting while
keeping function schemas in the task system prompt; Database uses a
fixed-schema delayed-query construction; and Code uses HumanEval-style
function-specification segmentation, conversational rewriting, and validation.
All compared methods use the same generated shard files.

\clearpage
\raggedbottom
\section{Qualitative Case Studies}
\label{sec:appendix-cases}

The cases below are real SHARDED traces. Each card shows the base rollout
followed by the \method{} rollout; user and assistant turns remain in their
original order. System prompts, verifier logs, timestamps, and cost metadata are
omitted. For prose-heavy responses, the cards report the verifier-observed
response type for non-final turns and the exact submitted answer, tool call, or
SQL artifact for turns scored as answer attempts. Markdown code fences are
removed only as typesetting wrappers.

\subsection{Successful Recoveries}
\label{sec:appendix-success-cases}

\promptheading{Case C1: Actions, updating tool arguments across turns}
\begingroup\small
\promptfield{Evaluation item}
\texttt{bfcl\_v3\_parallel\_multiple\_23}; strict tool-call AST match.
The available functions are \texttt{financial\_ratio.net\_profit\_margin} and
\texttt{financial\_ratio.debt\_ratio}.

\promptfield{Base rollout (\badtag{FAIL}): later numeric shards are not absorbed into the tool state}
\caseuser{1}{net profit margin Company XYZ}
\caseassistantblock{1}
\begin{lstlisting}[style=maigocase]
[financial_ratio.net_profit_margin(net_income=100000, total_revenue=500000)]
\end{lstlisting}
\caseuser{2}{The net income is \$20,000.}
\caseassistantblock{2}
\begin{lstlisting}[style=maigocase]
None of the provided functions match the given input format.
\end{lstlisting}
\caseuser{3}{The total revenue is \$100,000.}
\caseassistantblock{3}
\begin{lstlisting}[style=maigocase]
None of the provided functions match the given input format.
\end{lstlisting}
\caseuser{4}{Now calculate the debt ratio.}
\caseassistantblock{4}
\begin{lstlisting}[style=maigocase]
[financial_ratio.debt_ratio(total_liabilities=60000, total_assets=150000)]
\end{lstlisting}
\caseuser{5}{The total liabilities are \$10,000.}
\caseassistantblock{5}
\begin{lstlisting}[style=maigocase]
None of the provided functions match the given input format.
\end{lstlisting}
\caseuser{6}{The total assets are \$30,000.}
\caseassistantblock{6}
\begin{lstlisting}[style=maigocase]
None of the provided functions match the given input format.
\end{lstlisting}
\caseresult{\badtag{FAIL}}

\promptfield{\method{} rollout (\goodtag{PASS}): the final call list keeps both updated computations}
\caseuser{1}{net profit margin Company XYZ}
\caseassistantblock{1}
\begin{lstlisting}[style=maigocase]
[financial_ratio.net_profit_margin(net_income=100000, total_revenue=500000)]
\end{lstlisting}
\caseuser{2}{The net income is \$20,000}
\caseassistantblock{2}
\begin{lstlisting}[style=maigocase]
[financial_ratio.net_profit_margin(net_income=20000, total_revenue=500000)]
\end{lstlisting}
\caseuser{3}{The total revenue is \$100,000.}
\caseassistantblock{3}
\begin{lstlisting}[style=maigocase]
[financial_ratio.net_profit_margin(net_income=20000, total_revenue=100000)]
\end{lstlisting}
\caseuser{4}{Now calculate the debt ratio.}
\caseassistantblock{4}
\begin{lstlisting}[style=maigocase]
[financial_ratio.net_profit_margin(net_income=20000, total_revenue=100000),
 financial_ratio.debt_ratio(total_liabilities=50000, total_assets=150000)]
\end{lstlisting}
\caseuser{5}{The total liabilities are \$10,000.}
\caseassistantblock{5}
\begin{lstlisting}[style=maigocase]
[financial_ratio.net_profit_margin(net_income=20000, total_revenue=100000),
 financial_ratio.debt_ratio(total_liabilities=10000, total_assets=150000)]
\end{lstlisting}
\caseuser{6}{The total assets are \$30,000.}
\caseassistantblock{6}
\begin{lstlisting}[style=maigocase]
[financial_ratio.net_profit_margin(net_income=20000, total_revenue=100000),
 financial_ratio.debt_ratio(total_liabilities=10000, total_assets=30000)]
\end{lstlisting}
\caseresult{\goodtag{PASS}}

\promptfield{Takeaway}
\method{} still makes tentative intermediate calls, but unlike the base model
it continues revising the tool state as later numeric shards arrive and submits
the completed pair of computations at the final turn.
\par\endgroup\promptend

\promptheading{Case C2: Math, recovering the original question after later facts}
\begingroup\small
\promptfield{Evaluation item}
\texttt{gsm8k\_test/399}; exact numeric answer match.

\promptfield{Base rollout (\badtag{FAIL}): the final answer switches to the most recent relation}
\caseuser{1}{What is the difference between Alice's age and Erica's age?}
\caseassistantblock{1 (non-final hedge)}
\begin{lstlisting}[style=maigocase]
No submitted numeric answer.
\end{lstlisting}
\caseuser{2}{Erica is 30 years old.}
\caseassistantblock{2 (non-final hedge)}
\begin{lstlisting}[style=maigocase]
No submitted numeric answer.
\end{lstlisting}
\caseuser{3}{Beth is 5 years older than Erica.}
\caseassistantblock{3 (submitted answer)}
\begin{lstlisting}[style=maigocase]
5
\end{lstlisting}
\caseuser{4}{Alice is 7 years older than Beth.}
\caseassistantblock{4 (submitted answer)}
\begin{lstlisting}[style=maigocase]
7
\end{lstlisting}
\caseresult{\badtag{FAIL}}

\promptfield{\method{} rollout (\goodtag{PASS}): the final answer returns to Alice versus Erica}
\caseuser{1}{What is the difference between Alice's age and Erica's age?}
\caseassistantblock{1 (non-final hedge)}
\begin{lstlisting}[style=maigocase]
No submitted numeric answer.
\end{lstlisting}
\caseuser{2}{Erica is 30 years old.}
\caseassistantblock{2 (non-final hedge)}
\begin{lstlisting}[style=maigocase]
No submitted numeric answer.
\end{lstlisting}
\caseuser{3}{Beth is 5 years older than Erica.}
\caseassistantblock{3 (submitted answer)}
\begin{lstlisting}[style=maigocase]
5
\end{lstlisting}
\caseuser{4}{Alice is 7 years older than Beth.}
\caseassistantblock{4 (submitted answer)}
\begin{lstlisting}[style=maigocase]
12
\end{lstlisting}
\caseresult{\goodtag{PASS}}

\promptfield{Takeaway}
The base model solves the latest age relation but answers Alice versus Beth.
\method{} combines all revealed facts and preserves the original Alice-versus-Erica
query.
\par\endgroup\promptend

\promptheading{Case C3: Database, preserving an earlier SQL filter}
\begingroup\small
\promptfield{Evaluation item}
\texttt{spider\_dev/220}; execution-based SQL score.

\promptfield{Base rollout (\badtag{FAIL}): the final SQL drops the United Airlines filter}
\caseuser{1}{How many flights?}
\caseassistantblock{1 (submitted SQL)}
\begin{lstlisting}[style=maigocase]
SELECT COUNT(*) AS NumberOfFlights
FROM flights;
\end{lstlisting}
\caseuser{2}{Just United Airlines flights.}
\caseassistantblock{2 (submitted SQL)}
\begin{lstlisting}[style=maigocase]
SELECT COUNT(*) AS NumberOfUnitedFlights
FROM flights f
JOIN airlines a ON f.Airline = a.uid
WHERE a.Airline = 'United Airlines';
\end{lstlisting}
\caseuser{3}{Just the ones that arrive in Aberdeen.}
\caseassistantblock{3 (submitted SQL)}
\begin{lstlisting}[style=maigocase]
SELECT COUNT(*) AS NumberOfAberdeenArrivals
FROM flights f
JOIN airports a ON f.DestAirport = a.AirportCode
WHERE a.City = 'Aberdeen';
\end{lstlisting}
\caseuser{4}{Just the total count of those.}
\caseassistantblock{4 (submitted SQL)}
\begin{lstlisting}[style=maigocase]
SELECT COUNT(*) AS NumberOfAberdeenArrivals
FROM flights f
JOIN airports a ON f.DestAirport = a.AirportCode
WHERE a.City = 'Aberdeen';
\end{lstlisting}
\caseresult{\badtag{FAIL}}

\promptfield{\method{} rollout (\goodtag{PASS}): the final SQL keeps both constraints}
\caseuser{1}{How many flights?}
\caseassistantblock{1}
\begin{lstlisting}[style=maigocase]
Could you please specify which aspect of the flights you are interested in counting?
\end{lstlisting}
\caseuser{2}{Just United Airlines flights.}
\caseassistantblock{2 (submitted SQL)}
\begin{lstlisting}[style=maigocase]
SELECT COUNT(*) AS NumberOfUnitedFlights
FROM flights f
JOIN airlines a ON f.Airline = a.uid
WHERE a.Airline = 'United Airlines';
\end{lstlisting}
\caseuser{3}{Just the ones that arrive in Aberdeen.}
\caseassistantblock{3 (submitted SQL)}
\begin{lstlisting}[style=maigocase]
SELECT COUNT(*) AS NumberOfFlights
FROM flights f
JOIN airports a ON f.DestAirport = a.AirportCode
JOIN airlines al ON f.Airline = al.uid
WHERE a.City = 'Aberdeen';
\end{lstlisting}
\caseuser{4}{Yeah, I just want the total count of those flights.}
\caseassistantblock{4 (submitted SQL)}
\begin{lstlisting}[style=maigocase]
SELECT COUNT(*) AS NumberOfFlights
FROM flights
JOIN airports ON flights.DestAirport = airports.AirportCode
JOIN airlines ON flights.Airline = airlines.uid
WHERE airports.City = 'Aberdeen' AND airlines.Airline = 'United Airlines';
\end{lstlisting}
\caseresult{\goodtag{PASS}}

\promptfield{Takeaway}
The base model treats the Aberdeen shard as a replacement for the airline
constraint. \method{} recovers both filters when producing the final executable
query.
\par\endgroup\promptend

\subsection{A Remaining Failure Mode}
\label{sec:appendix-failure-case}

\promptheading{Case C4: A residual argument-binding error after requirement retention}
\begingroup\small
\promptfield{Task instance}
\texttt{bfcl\_v3\_parallel\_181}; strict tool-call AST match. The target is
four \texttt{get\_stock\_prices} calls: Apple for 1 week, Microsoft for 2
weeks, Amazon for 3 weeks, and Tesla for 1 month.

\promptfield{Base rollout (\badtag{FAIL}): later duration shards overwrite the earlier plan}
\caseuser{1}{get stock prices}
\caseassistantblock{1}
\begin{lstlisting}[style=maigocase]
[get_stock_prices(companies=['AAPL', 'GOOGL'], duration='1 week')]
\end{lstlisting}
\caseuser{2}{also for Apple}
\caseassistantblock{2}
\begin{lstlisting}[style=maigocase]
[get_stock_prices(companies=['AAPL'], duration='1 week')]
\end{lstlisting}
\caseuser{3}{and Microsoft}
\caseassistantblock{3}
\begin{lstlisting}[style=maigocase]
[get_stock_prices(companies=['AAPL', 'MSFT'], duration='1 week')]
\end{lstlisting}
\caseuser{4}{also Amazon}
\caseassistantblock{4}
\begin{lstlisting}[style=maigocase]
[get_stock_prices(companies=['AAPL', 'MSFT', 'AMZN'], duration='1 week')]
\end{lstlisting}
\caseuser{5}{and Tesla too}
\caseassistantblock{5}
\begin{lstlisting}[style=maigocase]
[get_stock_prices(companies=['AAPL', 'MSFT', 'AMZN', 'TSLA'], duration='1 week')]
\end{lstlisting}
\caseuser{6}{use the get\_stock\_prices function}
\caseassistantblock{6}
\begin{lstlisting}[style=maigocase]
[get_stock_prices(companies=['AAPL', 'MSFT', 'AMZN', 'TSLA'], duration='1 week')]
\end{lstlisting}
\caseuser{7}{for 1 week duration}
\caseassistantblock{7}
\begin{lstlisting}[style=maigocase]
[get_stock_prices(companies=['AAPL', 'MSFT', 'AMZN', 'TSLA'], duration='1 week')]
\end{lstlisting}
\caseuser{8}{for the next one, make it 2 weeks}
\caseassistantblock{8}
\begin{lstlisting}[style=maigocase]
[get_stock_prices(companies=['AAPL', 'MSFT', 'AMZN', 'TSLA'], duration='2 weeks')]
\end{lstlisting}
\caseuser{9}{and then 3 weeks after that}
\caseassistantblock{9}
\begin{lstlisting}[style=maigocase]
[get_stock_prices(companies=['AAPL', 'MSFT', 'AMZN', 'TSLA'], duration='3 weeks')]
\end{lstlisting}
\caseuser{10}{and for the last one, make it 1 month.}
\caseassistantblock{10}
\begin{lstlisting}[style=maigocase]
[get_stock_prices(companies=['AAPL', 'MSFT', 'AMZN', 'TSLA'], duration='1 month')]
\end{lstlisting}
\caseresult{\badtag{FAIL}}

\promptfield{\method{} rollout (\badtag{FAIL}): the duration sequence is retained, but binding remains over-broad}
\caseuser{1}{get stock prices}
\caseassistantblock{1}
\begin{lstlisting}[style=maigocase]
[get_stock_prices(companies=[], duration="1 week")]
\end{lstlisting}
\caseuser{2}{for Apple}
\caseassistantblock{2}
\begin{lstlisting}[style=maigocase]
[get_stock_prices(companies=['Apple'], duration="1 week")]
\end{lstlisting}
\caseuser{3}{and Microsoft}
\caseassistantblock{3}
\begin{lstlisting}[style=maigocase]
[get_stock_prices(companies=['Apple', 'Microsoft'], duration="1 week")]
\end{lstlisting}
\caseuser{4}{also Amazon}
\caseassistantblock{4}
\begin{lstlisting}[style=maigocase]
[get_stock_prices(companies=['Apple', 'Microsoft', 'Amazon'], duration="1 week")]
\end{lstlisting}
\caseuser{5}{and Tesla too}
\caseassistantblock{5}
\begin{lstlisting}[style=maigocase]
[get_stock_prices(companies=['Apple', 'Microsoft', 'Amazon', 'Tesla'],
                  duration="1 week")]
\end{lstlisting}
\caseuser{6}{use the get\_stock\_prices function}
\caseassistantblock{6}
\begin{lstlisting}[style=maigocase]
[get_stock_prices(companies=['Apple', 'Microsoft', 'Amazon', 'Tesla'],
                  duration="1 week")]
\end{lstlisting}
\caseuser{7}{for 1 week duration}
\caseassistantblock{7}
\begin{lstlisting}[style=maigocase]
[get_stock_prices(companies=['Apple', 'Microsoft', 'Amazon', 'Tesla'],
                  duration="1 week")]
\end{lstlisting}
\caseuser{8}{Actually, make it 2 weeks for the next one.}
\caseassistantblock{8}
\begin{lstlisting}[style=maigocase]
[get_stock_prices(companies=['Apple', 'Microsoft', 'Amazon', 'Tesla'],
                  duration="1 week"),
 get_stock_prices(companies=['Apple', 'Microsoft', 'Amazon', 'Tesla'],
                  duration="2 weeks")]
\end{lstlisting}
\caseuser{9}{and then 3 weeks after that}
\caseassistantblock{9}
\begin{lstlisting}[style=maigocase]
[get_stock_prices(companies=['Apple', 'Microsoft', 'Amazon', 'Tesla'],
                  duration="1 week"),
 get_stock_prices(companies=['Apple', 'Microsoft', 'Amazon', 'Tesla'],
                  duration="2 weeks"),
 get_stock_prices(companies=['Apple', 'Microsoft', 'Amazon', 'Tesla'],
                  duration="3 weeks")]
\end{lstlisting}
\caseuser{10}{And one month for the last one.}
\caseassistantblock{10}
\begin{lstlisting}[style=maigocase]
[get_stock_prices(companies=['Apple', 'Microsoft', 'Amazon', 'Tesla'],
                  duration="1 week"),
 get_stock_prices(companies=['Apple', 'Microsoft', 'Amazon', 'Tesla'],
                  duration="2 weeks"),
 get_stock_prices(companies=['Apple', 'Microsoft', 'Amazon', 'Tesla'],
                  duration="3 weeks"),
 get_stock_prices(companies=['Apple', 'Microsoft', 'Amazon', 'Tesla'],
                  duration="1 month")]
\end{lstlisting}
\caseresult{\badtag{FAIL}}

\promptfield{Takeaway}
\method{} recovers the multi-call structure and retains all four requested
durations. The remaining error is finer-grained: the model over-broadly binds
each duration to the accumulated company set rather than pairing one company
with one duration.
\par\endgroup\promptend

\end{document}